\pdfoutput=1

\documentclass[11pt]{article}

\usepackage[]{acl}
\usepackage{authblk}

\usepackage{times}
\usepackage{latexsym}
\usepackage{float}

\usepackage[T2A, T1]{fontenc}

\usepackage[utf8]{inputenc}

\usepackage{microtype}

\usepackage{inconsolata}

\usepackage{graphicx}

\usepackage{adjustbox}
\usepackage{array}
\usepackage{booktabs}
\usepackage{amsmath}
\usepackage{amsfonts}
\usepackage{multirow}
\usepackage{listings}
\usepackage{url}
\usepackage[ruled,linesnumbered]{algorithm2e}
\usepackage{CJKutf8}
\usepackage{subfigure}
\usepackage{supertabular}
\usepackage{dcolumn}
\usepackage{tikz}
\usepackage[right]{rotating}

\title{Breaking the Script Barrier in Multilingual Pre-Trained Language Models with Transliteration-Based Post-Training Alignment}

\author[1,*]{\bf Orgest Xhelili}
\author[2,3,*]{\bf Yihong Liu}
\author[2,3]{\bf Hinrich Sch\"utze}

\affil[1]{Technical University of Munich}
\affil[2]{Center for Information and Language Processing, LMU Munich} \affil[3]{Munich Center for Machine Learning (MCML)
\protect\\ \texttt{orgest.xhelili@tum.de, yihong@cis.lmu.de}}

\begin{document}
\maketitle
\def\thefootnote{*}\footnotetext{Equal contribution.}\def\thefootnote{\arabic{footnote}}

\def\secref#1{\S\ref{sec:#1}}
\def\seclabel#1{\label{sec:#1}}

\def\groupone{Mediterranean-Amharic-Farsi\xspace}
\def\grouptwo{South+East Asian Languages\xspace}

\newcommand{\udl}[1]{\underline{#1}}

\newcounter{notecounter}
\newcommand{\enotesoff}{\long\gdef\enote##1##2{}}
\newcommand{\enoteson}{\long\gdef\enote##1##2{{
\stepcounter{notecounter}
{\large\bf
\hspace{1cm}\arabic{notecounter} $<<<$ ##1: ##2
$>>>$\hspace{1cm}}}}}
\enoteson
\enotesoff

\begin{abstract}
Multilingual pre-trained models (mPLMs) have shown impressive performance on cross-lingual transfer tasks.
However, the transfer performance is often hindered when a low-resource target language is written in a different script than the
high-resource source language, even though the two languages may be related or share parts of their vocabularies.
Inspired by recent work that uses transliteration to address
this problem, our paper proposes a transliteration-based
post-pretraining alignment (PPA) method
aiming to improve the cross-lingual alignment between languages using diverse scripts.
We select two areal language groups, \textbf{\groupone} and \textbf{\grouptwo}, wherein the languages
are mutually influenced but use different scripts. We apply our method to these language groups and conduct extensive experiments
on a spectrum of downstream tasks. The results show that
after PPA, models  consistently outperform the original
model (up to 50\% for some tasks) in English-centric
transfer. In addition, when we use languages other
than English as sources in transfer,
our method obtains even larger improvements. We will make our code and models publicly available at  \url{https://github.com/cisnlp/Transliteration-PPA}.

\end{abstract}

\section{Introduction}\seclabel{sec:introduction}
Recent mPLMs such as mBERT~\citep{devlin-etal-2019-bert} and XLM-R~\citep{conneau-etal-2020-unsupervised}
have shown remarkable performance on cross-lingual transfer tasks by learning cross-lingual representations
from monolingual corpora~\citep{pires-etal-2019-multilingual, artetxe-etal-2020-cross}.
Despite their impressive performance, these models still exhibit limitations in cross-lingual transfer involving low-resource languages.
\citet{deshpande-etal-2022-bert} showed that the downstream performance of mPLMs is correlated with the degree of alignment between word embeddings across languages. Another factor that hinders the
knowledge transfer is the script diversity or script barrier of represented languages, which has been observed even in the case of
related languages~\citep{anastasopoulos-neubig-2019-pushing,muller-etal-2021-unseen}. The script barrier problem can also be viewed from
the perspective of representation alignment. \citet{wen-yi-mimno-2023-hyperpolyglot} showed that token representations from different scripts could be almost
perfectly linearly separated, indicating that models struggle to learn a common representation space.
Therefore, post-training is required to boost zero-shot cross-lingual transfer in tasks like sentence retrieval, text classification,
or sequence labeling, all of which benefit from better cross-lingual alignment~\citep{hammerl2024understanding}.

Many post-training alignment strategies use objectives that rely on bilingual dictionaries or parallel data to align the representations of mPLMs \citep{cao2019multilingual, wang2019cross, schuster-etal-2019-cross, pan-etal-2021-multilingual}.
However, dictionaries and parallel corpora are often limited in the data scale or the number of languages they
cover \citep{artetxe-etal-2020-call}, which might be impractical for building strong supervision signals for many low-resource languages.
Another alternative that improves cross-lingual alignment is to use \emph{transliteration} (a process of converting the text
of a language from one script to another). Transliteration can improve the lexical overlap, especially for related
languages \citep{moosa-etal-2023-transliteration}. Unlike translation, transliterations can be obtained nearly for free using
well-performing rule-based tools \citep{hermjakob-etal-2018-box}. Therefore, several works have shown improvements in cross-lingual
transfer by pre-training or fine-tuning models with data transliterated into a common script \citep{murikinati-etal-2020-transliteration,muller-etal-2021-unseen,purkayastha-etal-2023-romanization, moosa-etal-2023-transliteration}.
However, these methods require the model to use a single common script, which is restrictive for many tasks as the transliteration process can be lossy and non-invertible.

Recently, \citet{liu2024translico} proposed a sequence-level contrastive learning objective to improve the alignment across different scripts
at a large scale (for more than 500 languages), using both original script sentences and their Latin transliterations,
validated by English-centric cross-lingual zero-shot transfer evaluations.
However, there are three major limitations in their setup. First, the contrastive objective only manipulates the sequence-level representations
in the middle layer, which does not directly contribute to better alignment in the token-level space.
Second, not every language pair has extensive lexical overlap that can boost cross-lingual transfer: transliteration-based alignment
makes more sense for mutually influenced languages. Lastly, English alone as a transfer source language does not fully exploit the alignment
benefit, as it does not have the most lexical overlap with other languages.

To this end, we propose a new transliteration-based
post-training alignment method that works on
both \textbf{sequence} and \textbf{token} levels.
Our method does not rely on parallel data. Instead, similar to \citet{liu2024translico}, we use the monolingual data in their original scripts and
their Latin transliteration obtained by using \texttt{Uroman}~\citep{hermjakob-etal-2018-box}.
We investigate the impact of the strategy by focusing on two groups of languages: \textbf{\groupone} and \textbf{\grouptwo}, described
more in detail in Section~\ref{subsec:languages-and-data}.
The languages in each group share areal features but differ
in scripts. Some languages are closely related as members of the same
language family (e.g., Semitic and Sino-Tibetan). Additionally, languages in each group have extensive lexical overlap due to historical contact
and geographical proximity (e.g., Chinese and Korean or Turkish and Arabic). As these languages are written in different scripts, transliteration
can help to better exploit the shared linguistic properties and thus improve the cross-lingual transfer.

We leverage our method to post-train Glot500~\citep{imanigooghari-etal-2023-glot500} (a continually pre-trained model from XLM-R on more
than 500 languages) on the selected language groups and evaluate the zero-shot cross-lingual transfer performance on three types
of downstream tasks: sentence retrieval, text classification, and sequence labeling. The evaluation is done with English as the source
language and three other source languages of different scripts in each group. We show that our method consistently
improves the downstream task performance across different languages and scripts. Moreover, our method further boosts
the transfer performance when better source languages are chosen, as the performance depends on the degree of alignment between the source
and target languages -- precisely the alignment our approach boosts.

Our contributions can be summarized as follows: (i) We propose a transliteration-based post-training alignment method that operates on both
sequence  and token levels, aiming to bridge the script
barrier in mPLMs; (ii) We investigate the impact of our
method on two areal groups of
languages with different scripts and show consistent improvements in zero-shot cross-lingual transfer; (iii) We systematically explore how
different source languages influence the zero-shot transfer performance of our obtained transliteration-aligned models.

\section{Related Work}\seclabel{sec:related-work}
Many recent works have proposed pre-training or fine-tuning alignment methods to improve cross-lingual transfer in mPLMs.
\citet{cao2019multilingual} proposed a fine-tuning embedding alignment objective between word pairs procured in an unsupervised
fashion from parallel data using statistical word alignment models~\citep{dyer-etal-2013-simple}.
\citet{chaudhary2020dict} improved alignment during pre-training by using bilingual
dictionaries to replace words in original sentences with translations in other languages. Similarly, \citet{tang2022align}
used bilingual dictionaries to explicitly align the embeddings of the same words in different languages during pre-training.
\citet{wei2020learning} proposed a hierarchical contrastive learning pre-training method, which uses parallel data to align
representations at the word and sentence levels. Likewise, \citet{hu-etal-2021-explicit} proposed a pre-training method with explicit alignment
signals from parallel data that encourages symmetry at both word and sentence levels.
\citet{pan-etal-2021-multilingual} combined contrastive learning with translation language modeling~\citep{conneau2019cross} as a post-training
alignment method that uses parallel data as well. While these methods have shown improvements in cross-lingual transfer, they have
the limitation of requiring parallel data or bilingual dictionaries, which may be hard to acquire for many low-resource languages.

Transliteration is a process of converting text from one language script to another~\citep{wellisch1978conversion}.
This process does not involve translating meanings but rather represents the original symbols as closely as possible in the target script.
Different works have proposed transliteration-based methods to address the script barrier problem in multilingual models.
\citet{murikinati-etal-2020-transliteration} used transliteration to a common script to improve cross-lingual morphological inflection.
\citet{khemchandani-etal-2021-exploiting} exploited language relatedness between Indian languages and leveraged transliteration to a common script
to adapt multilingual models to low-resource languages.
\citet{muller-etal-2021-unseen} analyzed the behavior of multilingual models on unseen languages and found that languages written
in different scripts do not benefit from transfer learning.
They proposed transliteration to the high-resource source language script to address the script barrier.
\citet{purkayastha-etal-2023-romanization} showed that fine-tuning multilingual models on data transliterated into Latin script improves
cross-lingual transfer for low-resource languages. Similarly, \citet{moosa-etal-2023-transliteration} pre-trained models from scratch on data
transliterated into a common script for the Indic languages and showed improvements in cross-lingual transfer.
Our work is most related to \textsc{TransliCo} proposed by \citet{liu2024translico}, where a sequence-level contrastive learning objective is used
to encourage alignment across different scripts without restricting the models to a common script, using original script sentences and their Latin transliterations.
However, their English-centric evaluation setup limits the ability to fully reveal the impact of transliteration. This paper systematically explores how transliteration-based alignment enhances transfer performance by using various source languages.

\section{Methods}
\begin{figure*}[htb!]
\setlength{\belowcaptionskip}{-0.4cm}
\centering
\hspace{1cm}
\includegraphics[width=0.8\textwidth]{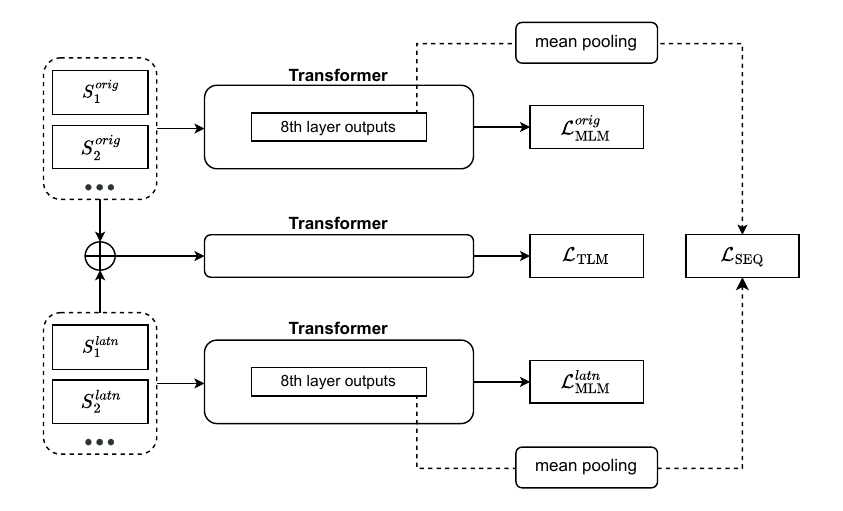}
\caption{
	Overview of our transliteration-based post-training alignment method consisting of three objectives: masked language modeling ($\mathcal{L}_\text{MLM}^{orig}$ and $ \mathcal{L}_\text{MLM}^{latn}$), sentence-level alignment ($\mathcal{L}_{\text{SEQ}}$), and token-level alignment ($\mathcal{L}_{\text{TLM}}$).
}
\label{fig:architecture}
\end{figure*}

\seclabel{sec:methods}
We present a transliteration-based post-training alignment method that can be used to fine-tune existing encoder-only mPLMs for improved alignment
across languages using different scripts, boosting cross-lingual transfer performance. Our method consists of three
objectives: masked language modeling, sentence-level alignment, and token-level alignment.
All objectives are trained on combined original and transliterated data. The transliterated data is obtained by converting the
original data into Latin script. The transliteration process uses \texttt{Uroman}~\citep{hermjakob-etal-2018-box}, a rule-based system that
can convert nearly all character sets into a common Latin script. The overall method is illustrated in Figure~\ref{fig:architecture}, and we
introduce the three objectives in detail below.

\subsection{Masked Language Modeling}\label{subsec:masked-language-modeling-(mlm)}
Given an input sequence in its original script: $X^{orig}_i$ or its transliterated version: $X^{latn}_i$,
we apply the naive MLM objective~\citep{devlin-etal-2019-bert} to predict randomly masked tokens in both sequences:
\begin{equation*}
    \mathcal{L}_{\text{MLM}} = \mathbb{E}\left[- \sum_{m \in \mathcal{M}} \log p_{\text{MLM}}(X_{i,m} | \boldsymbol{h}_{i,m})\right]
\end{equation*}
where $\mathcal{M}$ is the set of masked positions in the input sentence $X_i$ (either $X^{orig}_i$ or $X^{latn}_i$)
and $p_{\text{MLM}}(X_{i, m} | \boldsymbol{h}_{i,m})$ is the probability of predicting token $X_{i, m}$ given $\boldsymbol{h}_{i,m}$, the final
contextualized representation at the position $m$ in the $i$th sequence. The probability is computed by an MLM head. Fine-tuning with MLM on the original
data is necessary to preserve the model's knowledge. On the other hand, the mPLM has minimal knowledge about the transliterated data, which makes
the MLM objective on transliterated data crucial for learning useful cross-script representations. We refer to the MLM objective for the original
data (resp. transliterated data) as $ \mathcal{L}_\text{MLM}^{orig}$ (resp. $ \mathcal{L}_\text{MLM}^{latn}$).

\subsection{Sentence-Level Alignment}\seclabel{subsec:sentence-level-alignment}
We treat an input sequence in the original script $X^{orig}_i$ and its transliterated version $X^{latn}_i$ as having the same semantics.
Therefore, we apply a sequence-level contrastive learning objective, similar to SimCSE~\citep{gao-etal-2021-simcse},
to encourage the model to learn similar sequence-level representations for the original and transliterated sequences.
This setting is analogous to other works that apply contrastive learning on pairs formed by an original sentence and its English
translation~\citep{chi-etal-2021-infoxlm, pan-etal-2021-multilingual}. In our context, Latin acts as a pivot script, encouraging better
cross-lingual alignment of representations in different scripts.

Following \citet{liu2024translico}, we apply the contrastive learning objective on a given batch of original and transliterated sequences
$B = \{(X^{orig}_i, X^{latn}_i)\}_{i=1}^N$. Each batch defines positive contrastive pairs $(X, X^+)$ where $X$ is the original sequence
and $X^+$ is its transliterated version or vice versa, i.e., $(X^{orig}_i, X^{latn}_i)$ or $(X^{latn}_i, X^{orig}_i)$.
For each positive pair, the negative examples are formed by all other sequences in the batch
$B^- = B \setminus \{(X^{orig}_i, X^{latn}_i)\}$ (slightly abusing notation).
The contrastive loss is then defined as:
\begin{equation*}
\mathcal{L}_{\text{SEQ}} = \mathbb{E}\left[ -\log \frac{e^{\text{sim}(f(X), f(X^+))/\tau}}
{e^{\text{sim}(f(X), f(X^+))/\tau} + \text{NEG}}\right]\seclabel{eq:equation}
\end{equation*}
where $\text{NEG} = \sum_{(X,X^-) \in B^-} e^{\text{sim}(f(X), f(X^-))/\tau}$,
$f$ is defined as mean pooling over the 8th layer output contextualized embeddings (ignoring the special tokens' output except for [mask] token), $\text{sim}$ is the dot product,
and $\tau$ is the temperature set to $1$.

\subsection{Token-Level Alignment}\seclabel{subsec:token-level-alignment}
The sentence-level alignment objective helps the model to learn similar sentence representations for the original and transliterated sequences.
This is useful for improving performance in sentence-level downstream tasks like sentence retrieval or classification.
However, this objective manipulates the output of a middle layer, which does not directly contribute to better alignment in the token-level space.
For token-level tasks like NER and POS tagging, alignment at the token level might be more beneficial.
Therefore, we propose a token-level alignment objective that further encourages the model to align the representations of the original and
transliterated words. We adapt the translation language modeling objective introduced by ~\citet{conneau2019cross}, which is equivalent to applying
the MLM objective on a concatenated bilingual sentence pair. Specifically, given a sentence pair $(X^{orig}_i, X^{latn}_i)$, we apply the MLM objective
on the concatenated sequence $X^{orig}_i \oplus X^{latn}_i$ or $X^{latn}_i \oplus X^{orig}_i$, where the concatenation order is randomly chosen
during training.
The intuition is that, to predict a token masked in the original sentence, the model can either attend to surrounding tokens in the original script or their transliterations and vice versa.
This encourages the model to align the representations in the original and the Latin script.
We refer to this objective as \textbf{transliteration language modeling} (TLM) and the loss as $\mathcal{L}_{\text{TLM}}$.

The overall training objective combines the masked language modeling, sentence-level alignment, and token-level alignment objectives:
\begin{equation*}
\mathcal{L} = \mathcal{L}_\text{MLM}^{orig} + \mathcal{L}_\text{MLM}^{latn} + \mathcal{L}_{\text{SEQ}} + \mathcal{L}_{\text{TLM}}\seclabel{eq:equation2}
\end{equation*}

\section{Experiments}\seclabel{sec:experiments}
\subsection{General Setups}\label{subsec:general-setups}
We use the Glot500 model~\citep{imanigooghari-etal-2023-glot500}, a state-of-the-art multilingual encoder-only model
pre-trained on more than 500 languages, as our source model for all our experiments. We fine-tune Glot500 on two groups of languages using
the proposed transliteration-based post-training alignment method. The languages for each group are selected based on \textbf{areal features} so
that they have some lexical overlap in different degrees and cover various scripts.
We then evaluate the two resulting models on several downstream tasks in a zero-shot cross-lingual transfer manner.
Apart from the standard transfer setting with English as the source language,
we also evaluate the model's transfer capabilities with three other source languages of different scripts for each language group.

\begin{figure*}[htb!]
\setlength{\belowcaptionskip}{-0.4cm}
\centering
\includegraphics[width=1\textwidth]{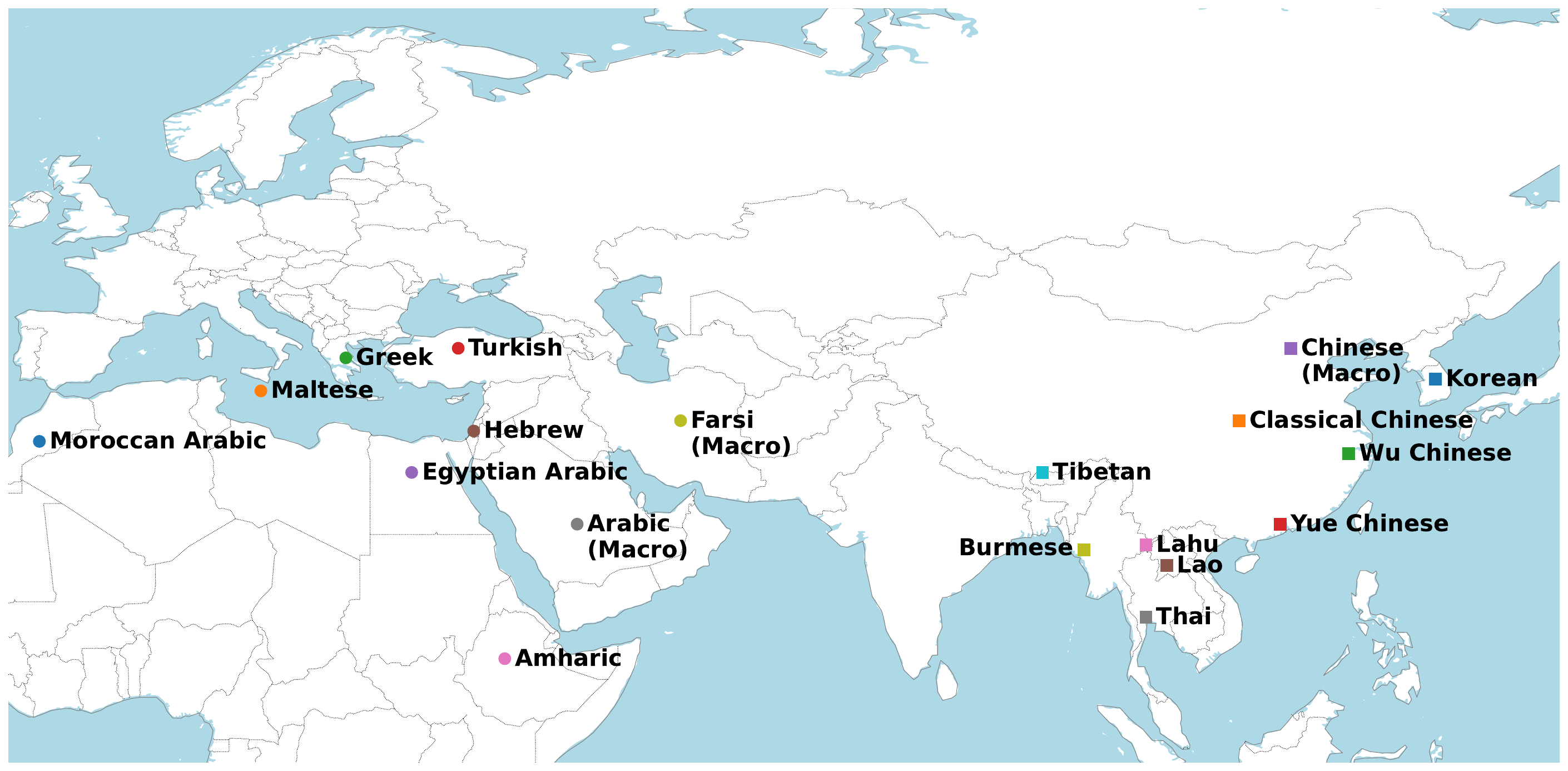}
\caption{Geographical distribution of languages selected in each group. \textbf{\groupone} are shown with circles, while
\textbf{\grouptwo} are shown with squares.}
\label{fig:map}
\end{figure*}

\subsection{Languages, Data and Models}\label{subsec:languages-and-data}
The two language groups are named as \textbf{\groupone} and \textbf{\grouptwo}. We visualize each group's geographical distribution
of the selected languages in Figure~\ref{fig:map}. The
languages within each group are spoken in adjacent areas, and
there is a long history of linguistic influence between
them. For example, Arabic has had extensive contact with
languages such as Turkish and Persian~\citep{versteegh2001linguistic}.
The data for each language is sampled from the Glot500-c training dataset~\citep{imanigooghari-etal-2023-glot500}.
We sample 10\% of the available data for each language, or a minimum of 10k sentences, whichever is larger.
The data is then transliterated into Latin script using \texttt{Uroman}~\citep{hermjakob-etal-2018-box}. Table~\ref{tab:languages} shows
each group's languages and the number of sampled sentences. In total, \groupone consists of 10 languages,
5 scripts, and around 16M sentences, while \grouptwo consists of 10 languages, 7 scripts, and around 4M sentences.
We fine-tune Glot500 using our alignment method on each group separately. We then select the best checkpoint for each group by validating
the checkpoints' performance on the Tatoeba~\citep{artetxe-schwenk-2019-massively} sentence retrieval dataset, which contains 1000 English-aligned sentences. We compute the top-10 retrieval accuracy
based on the cosine similarity of the averaged 8th-layer contextual embeddings. The best checkpoint for each group is regarded as our final aligned model.

\begin{table}[hbt!]
\setlength{\belowcaptionskip}{-0.5cm}
\centering
\small
\begin{adjustbox}{max width=.45\textwidth}
\begin{tabular}{lrrr}
\toprule
\textbf{Language} & \textbf{Script Code} & \textbf{Language Code} & \textbf{Num. Sent.} \\
\midrule
\multicolumn{4}{c}{\textbf{\groupone}} \\
\midrule
Macro Lang. Arabic & Arab & ara & 2.4M \\
Standard Arabic & Arab & arb & 15k \\
Moroccan Arabic & Arab & ary & 10k \\
Egyptian Arabic & Arab & arz & 348k \\
Macro Lang. Farsi & Arab & fas & 1.8M \\
Amharic & Ethi & amh & 286k \\
Greek & Grek & ell & 2.2M \\
Hebrew & Hebr & heb & 1.8M \\
Turkish & Latn & tur & 2.9M \\
Maltese & Latn & mlt & 4M \\
\midrule
\multicolumn{4}{c}{\textbf{\grouptwo}} \\
\midrule
Macro Lang. Chinese & Hani & zho & 2.4M \\
Classical Chinese & Hani & lzh & 10k \\
Yue Chinese & Hani & yue & 48k \\
Wu Chinese & Hani & wuu & 22k \\
Korean & Hang & kor & 646k \\
Lao & Laoo & lao & 10k \\
Lahu & Latn & lhu & 10k \\
Burmese & Mymr & mya & 94k \\
Tibetan & Tibt & bod & 27k \\
Thai & Thai & tha & 773k \\
\bottomrule
\end{tabular}
\end{adjustbox}
\caption{Basic information and number of sampled sentences for each language in the two groups.}
\label{tab:languages}
\end{table}

\begin{table*}[htb!]
\setlength{\belowcaptionskip}{-0.5cm}
\scriptsize
\setlength{\tabcolsep}{3pt}
\renewcommand{\arraystretch}{1.2} 
\centering
\begin{tabular}{l|cccc|cccc|cccc|cccc|cccc}
\toprule
& \multicolumn{4}{c|}{SR-B} & \multicolumn{4}{c|}{Taxi1500} & \multicolumn{4}{c|}{SIB200} & \multicolumn{4}{c|}{NER} & \multicolumn{4}{c}{POS} \\
& Latn & Arab & Grek & Hebr & Latn & Arab & Grek & Hebr & Latn & Arab & Grek & Hebr & Latn & Arab & Grek & Hebr & Latn & Arab & Grek & Hebr \\
\midrule
\multicolumn{21}{c}{\textbf{Glot500}} \\
\midrule
tur\_Latn & 63.2 & \udl{77.6} & 51.8 & 32.2 & 63.0 & \udl{63.5} & \textbf{55.6} & \textbf{44.1} & 81.4 & 79.8 & \udl{82.1} & 79.8 & \udl{74.1} & 71.4 & 71.7 & 70.4 & \udl{70.4} & 49.4 & 63.8 & 66.5 \\
mlt\_Latn & 50.4 & 55.0 & \udl{69.2} & 44.8 & 54.1 & \udl{57.3} & 53.0 & \textbf{46.1} & 81.8 & 79.3 & \udl{82.8} & 82.6 & \udl{69.2} & 60.0 & 68.2 & 66.9 & \udl{81.1} & 59.8 & 76.9 & 74.2 \\
ell\_Grek & 48.6 & \udl{58.6} & src  & 40.8 & \udl{\textbf{64.0}} & \textbf{61.4} & src  & \textbf{44.1} & 77.7 & 74.4 &  src & \udl{81.7} & \udl{72.7} & 72.1 &  src & 72.6 & \udl{\textbf{86.1}} & 58.3 &  src & 67.8 \\
heb\_Hebr & 21.8 & 27.8 & \udl{33.8} & src  & 35.4 & \udl{\textbf{44.9}} & 39.0 & src  & 77.6 & 73.9 & \udl{\textbf{81.7}} &  src & 48.9 & \udl{58.7} & 52.6 &  src & \textbf{68.3} & \udl{70.1} & \textbf{60.6} & src \\
amh\_Ethi & 52.8 & \udl{64.6} & 51.2 & 33.2 & 7.2  & \textbf{10.4} & 12.7 & \udl{\textbf{15.1}} & 73.1 & \udl{74.9} & 74.1 & 74.8 & 43.8 & \textbf{52.5} & \udl{54.0} & 46.6 & \textbf{66.5} & \textbf{65.1} & \textbf{64.3} & \udl{73.6} \\
ara\_Arab &  -   &  -   &   -  &   -  &  -   &  -   &  -   &  -   &  -   &  -   &    - &    - & 57.2 & src  & 56.7 & \udl{61.5} & \udl{\textbf{84.6}} &  src & \textbf{63.2} & \textbf{78.0} \\
arz\_Arab & 24.8 & 33.6 & \udl{52.8} & 44.8 & 35.1 & 43.1 & \udl{45.2} & \textbf{42.7} & 79.7 & src  & \udl{81.8} & 80.4 & \textbf{58.4} & \udl{75.1} & 63.8 & 65.2 &    - &    - &    - & - \\
ary\_Arab & 15.2 & 16.4 & \udl{29.0} & 28.4 & 35.8 & \textbf{40.3} & \udl{\textbf{41.6}} & 39.6 & 79.9 & 80.2 & \udl{\textbf{84.0}} & 82.0 &    - &    - &    - &    - &    - &    - &    - & - \\
arb\_Arab & 14.6 & 23.0 & 29.0 & \udl{32.2} &  -   &  -   &  -   &  -   & 79.9 & 79.9 & \udl{82.8} & 81.2 & -    & -    &    - &    - &    - &    - &    - & - \\
fas\_Arab & \udl{89.2} & src  & 72.4 & 40.2 & \udl{71.0} & src  & 59.2 & \textbf{48.7} &  -   &  -   &    - &    - & \textbf{49.7} & \udl{\textbf{66.2}} & 58.2 & 50.6 & 71.5 & 67.2 & 60.9 & \udl{\textbf{72.0}} \\
\midrule
\textbf{Average} & 42.2 & 44.5 & \udl{48.6} & 37.0 & 45.7 & \udl{45.8} & 43.8 & \textbf{40.1} & 78.9 & 77.5 & \udl{81.3} & 80.4 & 59.3 & \udl{65.1} & 60.8 & 62.0 & \udl{72.8} & 61.6 & 65.0 & 72.0 \\
\midrule
\multicolumn{21}{c}{\textbf{Ours}} \\
\midrule
tur\_Latn & \textbf{81.0} & \udl{\textbf{91.2}} & \textbf{77.8} & \textbf{49.4} & \textbf{64.8} & \udl{\textbf{65.1}} & 54.6 & 38.2 & \textbf{85.6} & \udl{\textbf{86.0}} & \textbf{84.8} & \textbf{85.6} & \udl{\textbf{77.0}} & \textbf{73.1} & \textbf{76.9} & \textbf{73.2} & \udl{\textbf{73.6}} & \textbf{52.8} & \textbf{66.4} & \textbf{68.0} \\
mlt\_Latn & \textbf{85.6} & \udl{\textbf{93.4}} & \textbf{90.4} & \textbf{59.4} & \udl{\textbf{66.6}} & \textbf{60.9} & \textbf{58.4} & 42.5 & \udl{\textbf{86.2}} & \textbf{85.8} & \textbf{84.6} & \textbf{85.4} & \udl{\textbf{75.2}} & \textbf{72.0} & \textbf{73.9} & \udl{\textbf{75.2}} & \udl{\textbf{83.1}} & \textbf{63.0} & \textbf{77.1} & \textbf{77.8} \\
ell\_Grek & \textbf{68.0} & \udl{\textbf{85.4}} & src  & \textbf{45.2} & \udl{63.6} & 60.4 & src  & 36.4 & \udl{\textbf{82.3}} & \textbf{81.4} & src  & \textbf{82.1} & \textbf{73.8} & \textbf{73.9} & src  & \udl{\textbf{75.3}} & \udl{85.9} & \textbf{58.7} & src  & \textbf{70.7} \\
heb\_Hebr & \textbf{29.0} & \textbf{32.0} & \udl{\textbf{42.8}} & src  & \textbf{45.3} & 44.5 & \udl{\textbf{45.8}} & src  & \textbf{79.0} & \textbf{79.8} & \udl{79.9} & src  & \textbf{51.9} & \udl{\textbf{61.6}} & \textbf{57.2} & src  & 67.7 & \udl{\textbf{71.3}} & 59.8 & src \\
amh\_Ethi & \textbf{63.6} & \udl{\textbf{79.4}} & \textbf{64.8} & \textbf{49.4} & \textbf{7.6 } & 9.6  & \udl{\textbf{17.0}} & 11.3 & \textbf{77.2} & \udl{\textbf{77.9}} & \textbf{76.6} & \textbf{76.9} & \textbf{45.0} & 51.7 & \udl{54.0} & \textbf{50.0} & 66.2 & 63.9 & 63.7 & \udl{\textbf{74.9}} \\
ara\_Arab & -    &   -  & -    & -    & -    & -    & -    & -    & -    & -    & -    & -    & \textbf{59.9} & src  & \textbf{60.4} & \udl{\textbf{65.1}} & 65.3 &  src & 62.3 & \udl{77.7} \\
arz\_Arab & \textbf{56.4} & \textbf{79.4} & \udl{\textbf{82.2}} & \textbf{69.6} & \textbf{41.8} & \textbf{44.1} & \udl{\textbf{49.6}} & 42.4 & \textbf{83.1} & src  & \udl{\textbf{83.4}} & \textbf{82.8} & 58.3 & \udl{\textbf{76.7}} & \textbf{65.6} & \textbf{68.3} & -    & -    & -    & - \\
ary\_Arab & \textbf{47.6} & \udl{\textbf{66.2}} & \textbf{66.2} & \textbf{65.4} & \textbf{39.0} & 37.3 & 39.0 & \udl{\textbf{40.5}} & \textbf{83.2} & \textbf{82.7} & 82.9 & \udl{\textbf{83.3}} & -    & -    & -    & -    & -    & -    & -    & - \\
arb\_Arab & \textbf{44.4} & \textbf{55.0} & \udl{\textbf{56.0}} & \textbf{49.6} & -    & -    & -    & -    & \textbf{82.8} & \udl{\textbf{83.3}} & \textbf{83.1} & \textbf{83.1} & -    & -    & -    & -    & -    & -    & -    & - \\
fas\_Arab & \udl{\textbf{89.6}} & src  & \textbf{87.0} & \textbf{57.8} & \udl{\textbf{71.9}} & src  & \textbf{63.2} & 37.5 & -    & -    & -    & -    & 48.6 & \udl{63.1} & \textbf{62.3} & \textbf{56.9} & \udl{\textbf{71.6}} & \textbf{69.1} & \textbf{61.5} & 71.3 \\
\midrule
\textbf{Average} & \textbf{62.8} & \udl{\textbf{72.7}} & \textbf{70.9} & \textbf{55.7} & \udl{\textbf{50.1}} & \textbf{46.0} & \textbf{46.8} & 35.5 &
\textbf{82.4} & \textbf{82.4} & \textbf{82.2} & \udl{\textbf{82.7}} & \textbf{61.2} & \udl{\textbf{67.4}} & \textbf{64.3} & \textbf{66.3} &
\textbf{73.3} & \textbf{63.1} & \textbf{65.1} & \udl{\textbf{73.4}} \\
\hline
\end{tabular}
\caption{Cross-lingual transfer results across 5 tasks on the \textbf{\groupone} group.
Columns represent the script of the source language (denoted with ``src''), while rows represent the target languages.
Results are averaged over 5 random seeds.
For each source-target language pair, the best score is \textbf{bolded}. For each target language, we \udl{underline} the best source transfer score
for each task (for both Glot500 and our method).}
\label{table:combined_group1}
\end{table*}

\subsection{Downstream Tasks}\label{subsec:downstream-tasks}
We evaluate the resulting aligned model for each group on several downstream tasks (described below).
For each task, we use \textbf{four} different source languages, \textbf{English} and \textbf{three other source languages} belonging to the same
group that use different scripts. The evaluation is done in a zero-shot cross-lingual transfer manner: we fine-tune the models on
the train set of a given source language, select the best
checkpoint, and compute the macro F1 score (except for SR-B
where we compute the average top-10 retrieval accuracy)
on the test sets of the remaining languages in each group. Note that no training step is needed for the retrieval task: we directly use the sentences from the source language as the
queries and retrieve the most similar sentences in the target languages. For tasks that require additional fine-tuning, we report the results averaged over five different seeds.
The downstream tasks are as follows (see details in \secref{sec:downstream-tasks}):

\paragraph{SR-B} A sentence retrieval dataset where the parallel sentences are from the Bible. We compute the top-10 retrieval accuracy on 500 parallel sentences following~\citet{imanigooghari-etal-2023-glot500}.

\paragraph{Taxi1500} A multilingual text classification dataset covering more than 1500 languages with sentences from 6 topics~\citep{ma2023taxi1500}.

\paragraph{SIB200} A multilingual text classification dataset covering more than 200 languages for 7 topics~\citep{adelani-etal-2024-sib}.

\paragraph{NER} A multilingual sequence labeling dataset for named entity recognition ~\citep{pan-etal-2017-cross} that consists of articles annotated with 7 different tags, e.g., location, person, etc.

\paragraph{POS} A multilingual sequence labeling dataset for part-of-speech (POS) tagging~\citep{de-marneffe-etal-2021-universal} consisting of sentences annotated with 17 universal POS tags, e.g., NOUN, ADJ, etc.

\begin{table*}[htb!]
\setlength{\belowcaptionskip}{-0.5cm}
\scriptsize
\setlength{\tabcolsep}{3.1pt}
\renewcommand{\arraystretch}{1.2} 
\centering
\begin{tabular}{l|cccc|cccc|cccc|cccc|ccc}
\toprule
& \multicolumn{4}{c|}{SR-B} & \multicolumn{4}{c|}{Taxi1500} & \multicolumn{4}{c|}{SIB200} & \multicolumn{4}{c|}{NER} & \multicolumn{3}{c}{POS} \\
& Latn & Hang & Hani & Thai & Latn & Hang & Hani & Thai & Latn & Hang & Hani & Thai & Latn & Hang & Hani & Thai & Latn & Hang & Hani \\
\midrule
\multicolumn{20}{c}{\textbf{Glot500}} \\
\midrule
tha\_Thai & \textbf{45.4} & \udl{47.2} & 42.6 & src & 64.1 & 63.8 & \udl{\textbf{73.6}} & src  & 82.0 & \udl{83.1} & \udl{\textbf{83.1}} & src  & \textbf{4.3 } & \textbf{2.8 } & \udl{7.5}  & src  & \udl{\textbf{55.0}} & 29.8 & \textbf{49.0} \\
kor\_Hang & \textbf{61.0} & src  & \udl{64.6} & 51.2 & \udl{68.6} & src  & 63.5 & 65.5 & 82.7 & src  & \udl{\textbf{83.0}} & 82.5 & \udl{51.8} & src  & 40.2 & \textbf{10.1} & \udl{52.6} & src  & 39.2 \\
yue\_Hani & 24.0 & 31.6 & \udl{65.8} & 44.4 & 64.0 & 62.8 & \udl{68.0} & 62.1 & 84.7 & 84.4 & src  & \udl{86.9} & \textbf{24.0} & \textbf{37.4} & \udl{69.1} & \textbf{16.4} & \textbf{38.8} & \textbf{49.3} & \udl{78.5} \\
wuu\_Hani & -    & -    & -    & -    & -    & -    & -    & -    & -    & -    & -    & -    & 35.1 & \textbf{58.3} & \udl{62.4} & \textbf{16.1} & -    & -    & -  \\
zho\_Hani & \textbf{44.4} & \udl{46.4} & src  & 39.6 & \udl{65.0} & 61.9 & src  & 62.0 & -    & -    & -    & -    & \textbf{23.6} & \udl{32.7} & src  & 15.0 & \textbf{40.1} & \udl{\textbf{49.9}} & src \\
lzh\_Hani & \textbf{63.4} & \textbf{65.8} & \udl{\textbf{75.8}} & \textbf{43.2} & 57.3 & \textbf{60.5} & \udl{\textbf{61.5}} & \textbf{56.1} & -    & -    & -    & -    & \textbf{12.0} & 29.8 & \udl{60.1} & \textbf{22.3} & \textbf{19.4} & \textbf{27.2} & \udl{50.7} \\
lao\_Laoo & 49.6 & 59.8 & 48.6 & \udl{64.6} & \textbf{72.0} & 68.5 & \textbf{73.9} & \udl{\textbf{74.8}} & 80.4 & 80.0 & 80.0 & \udl{82.4} & -    & -    & -    & -    & -    & -    & - \\
lhu\_Latn & 5.0  & 6.0  & 6.6  & \udl{7.2}  & \textbf{27.0} & \udl{34.1} & \textbf{27.8} & 26.4 & -    & -    & -    & -    & -    & -    & -    & -    & -    & -    & - \\
mya\_Mymr & 29.4 & \udl{37.8} & 29.0 & 33.6 & 61.8 & \udl{63.2} & 56.8 & 60.5 & 80.1 & \udl{80.7} & 78.6 & 79.5 & 54.1 & \udl{\textbf{65.2}} & \textbf{49.7} & \textbf{9.3 } & - & - & - \\
bod\_Tibt & 33.2 & 49.4 & 44.4 & \udl{49.8} & -    & -    & -    & -    & 70.0 & 68.8 & 65.1 & \udl{\textbf{72.7}} & 36.5 & 42.8 & \udl{\textbf{50.0}} & \textbf{25.5} & - & - & - \\
\midrule
\textbf{Average} & 39.4 & 43.0 & \udl{47.1} & 41.7 & 60.0 & 60.7 & \udl{\textbf{62.1}} & 59.9 & 79.3 & 79.4 & 78.0 & \udl{80.8} & 30.2 & 38.4 &
\udl{\textbf{48.4}} & \textbf{16.4} & \textbf{41.2} & \textbf{39.1} & \udl{54.4} \\
\midrule
\multicolumn{20}{c}{\textbf{Ours}} \\
\midrule
tha\_Thai & 45.2 & \udl{\textbf{85.6}} & \textbf{55.2} & src  & \textbf{66.3} & \textbf{66.7} & \udl{67.8} & src  & \udl{\textbf{86.6}} & \textbf{85.9} & 82.9 & src  & 3.5  & 2.6  & \udl{\textbf{8.6 }} & src  & \udl{50.6} & \textbf{30.3} & 48.6 \\
kor\_Hang & 58.8 & src  & \udl{\textbf{79.6}} & \textbf{76.6} & \udl{\textbf{71.3}} & src  & \textbf{65.8} & \textbf{68.2} & \textbf{83.1} & src  & 82.4 & \udl{\textbf{83.8}} & \udl{\textbf{55.5}} & src  & \textbf{43.2} & 3.1  & \udl{\textbf{52.8}} & src  & \textbf{45.0} \\
yue\_Hani & \textbf{71.4} & \textbf{91.8} & \udl{\textbf{98.8}} & \textbf{89.8} & \textbf{64.5} & \textbf{66.4} & \udl{\textbf{69.2}} & \textbf{67.9} & \textbf{87.2} & \textbf{84.7} & src  & \udl{\textbf{89.0}} & 21.0 & 36.1 & \udl{\textbf{70.1}} & 15.8 & 29.3 & 38.6 & \udl{\textbf{79.3}} \\
wuu\_Hani & -    & -    & -    & -    & -    & -    & -    & -    & -    & -    & -    & -    & \textbf{45.5} & 56.7 & \udl{\textbf{64.7}} & 1.7  & -    & -    & - \\
zho\_Hani & 41.4 & \udl{\textbf{75.0}} & src  & \textbf{46.4} & \textbf{65.8} & \textbf{65.0} & src  & \udl{\textbf{67.7}} & -    & -    & -    & -    & 21.5 & \udl{\textbf{35.9}} & src  & \textbf{15.6} & 32.2 & \udl{44.1} & src \\
lzh\_Hani & 37.4 & 48.6 & \udl{67.6} & 37.0 & \udl{\textbf{63.4}} & 58.3 & 61.4 & 55.2 & -    & -    & -    & -    & 11.7 & \textbf{33.7} & \udl{\textbf{61.4}} & 20.0 & 13.9 & 14.3 & \udl{\textbf{52.1}} \\
lao\_Laoo & \textbf{56.4} & \udl{\textbf{90.8}} & \textbf{67.0} & \textbf{77.4} & \udl{70.2} & \textbf{69.5} & 69.4 & 67.2 & \textbf{82.6} & \textbf{81.6} & \textbf{81.0} & \udl{\textbf{83.2}} & -    & -    & -    & -    & -    & -    & - \\
lhu\_Latn & \textbf{15.8} & \textbf{26.6} & \textbf{21.0} & \udl{\textbf{27.4}} & 23.9 & \udl{\textbf{36.5}} & 24.9 & \textbf{32.1} & -    & -    & -    & -    & -    & -    & -    & -    & -    & -    & - \\
mya\_Mymr & \textbf{37.8} & \udl{\textbf{60.4}} & \textbf{54.4} & \textbf{59.6} & \textbf{62.2} & \udl{\textbf{68.3}} & \textbf{59.2} & \textbf{63.0} & \textbf{80.9} & \udl{\textbf{81.8}} & \textbf{80.0} & \textbf{81.0} & \textbf{56.0} & \udl{62.4} & 45.0 & 5.8  & - & - & - \\
bod\_Tibt & \textbf{60.8} & \udl{\textbf{88.8}} & \textbf{83.4} & \textbf{80.2} & -    & -    & -    & -    & \textbf{70.5} & \udl{\textbf{73.5}} & \textbf{71.4} & 71.4 & \textbf{40.3} & \udl{\textbf{43.1}} & 41.6 & 1.1  & - & - & - \\
\midrule
\textbf{Average} & \textbf{47.2} & \udl{\textbf{70.9}} & \textbf{65.8} & \textbf{61.8} & \textbf{61.0} & \udl{\textbf{61.5}} & 59.7 &
\textbf{60.2} & \udl{\textbf{81.8}} & \textbf{81.5} & \textbf{79.6} & \textbf{81.7} & \textbf{31.9} & \textbf{38.6} & \udl{47.8} & 9.0 & 35.8 & 31.8 & \udl{\textbf{56.3}} \\
\hline
\end{tabular}
\caption{Cross-lingual transfer results across 5 tasks on the \textbf{\grouptwo} group.
Columns represent the script of the source language (denoted with ``src''), while rows represent the target languages.
Results are averaged over 5 random seeds.
For each source-target language pair, the best score is \textbf{bolded}. For each target language, we \udl{underline} the best source transfer score
for each task (for both Glot500 and our method)}
\label{table:combined_group2}
\end{table*}

\section{Results and Analysis}\seclabel{sec:results}
We report the results of Glot500 and our post-trained aligned models on the downstream tasks in Table~\ref{table:combined_group1} for \groupone and
in Table~\ref{table:combined_group2} for \grouptwo. Overall, our aligned models outperform Glot500 across different tasks
for both language groups, occasionally with a slight performance drop for certain source-target language combinations.
In the following, we highlight our essential findings from the results.

\paragraph{Per-group performances differ slightly.}
Starting with \textbf{\groupone}, we observe that the post-trained aligned model generally outperforms Glot500 on all downstream tasks.
The SR-B task shows the most significant improvement, with the aligned model achieving, on average, more than 20\% higher accuracy than Glot500
for all source languages. For other tasks, the aligned model also demonstrates a consistent, albeit smaller, improvement, with
Glot500 occasionally outperforming the post-trained aligned model for specific source-target language pairs.
However, we observe a more mixed performance for the \textbf{\grouptwo}, especially for the sequence labeling tasks, with the aligned models performing worse than Glot500 when transferring from more than half of the source languages. We hypothesize that this performance
drop is primarily due to the transliteration process, which loses the semantic and contextual nuances and induces more token-level ambiguity
for most of the languages~\citep{amrhein-sennrich-2020-romanization,liu2024translico}. This ambiguity makes token-level alignment more difficult.
In contrast, the aligned model achieves consistent improvements in NER and POS for \groupone, where less token-level ambiguity is introduced
as the languages are originally written in phonetic scripts, to which Latin also belongs.

\begin{table*}[ht]
\setlength{\belowcaptionskip}{-0.1cm}
\scriptsize
\setlength{\tabcolsep}{3pt}
\renewcommand{\arraystretch}{1.2} 
\centering
\begin{tabular}{l|cccc|cccc|cccc|cccc|cccc|}
\toprule
& \multicolumn{4}{c|}{SR-B} & \multicolumn{4}{c|}{Taxi1500} & \multicolumn{4}{c|}{SIB200} & \multicolumn{4}{c|}{NER} & \multicolumn{4}{c|}{POS} \\
& Latn & Arab & Grek & Hebr & Latn & Arab & Grek & Hebr & Latn & Arab & Grek & Hebr & Latn & Arab & Grek & Hebr & Latn & Arab & Grek & Hebr \\
\midrule
Glot500        & 42.2 & 44.5 & 48.6 & 37.0 & 45.7 & 45.8 & 43.8 & \textbf{40.1} & 78.9 & 77.5 & 81.3 & 80.4 & 59.3 & 65.1 & 60.8 & 62.0 & 72.8 & 61.6 & \udl{65.0} & 72.0 \\
\midrule
MLM            & 50.2 & 53.2 & 55.4 & 40.2 & 47.7 & 43.8 & 46.9 & 31.8 & \textbf{82.6} & \textbf{82.4} & \udl{82.5} & 80.4 & \udl{61.7} & \textbf{68.5} & 63.3 & \udl{66.5} & 71.9 & \udl{62.4} & 63.7 & 72.0 \\
MLM+SEQ        & \textbf{62.9} & \udl{71.9} & \udl{69.8} & \textbf{57.1} & 49.0 & \udl{46.0} & \textbf{48.5} & 36.1 & \udl{82.4} & \textbf{82.4} & \textbf{83.5} & \udl{82.4} & 61.1 & \udl{67.4} & \textbf{64.6} & 66.1 & 72.5 & 62.1 & 64.7 & 72.7 \\
MLM+TLM        & 50.4 & 54.7 & 55.7 & 41.6 & \textbf{50.4} & \textbf{46.5} & \udl{47.3} & \udl{37.2} & \textbf{82.6} & \udl{81.7} & 82.4 & 81.8 & \textbf{61.9} & 67.1 & 64.0 & \textbf{66.8} & \udl{73.2} & \textbf{63.1} & 64.5 & \textbf{73.5} \\
MLM+SEQ+TLM    & \udl{62.8} & \textbf{72.7} & \textbf{70.9} & \udl{55.7} & \udl{50.1} & \udl{46.0} & 46.8 & 35.5 & \udl{82.4} & \textbf{82.4} & 82.2 & \textbf{82.7} & 61.2 & \udl{67.4} & \udl{64.3} & 66.3 & \textbf{73.3} & \textbf{63.1} & \textbf{65.1} & \udl{73.4} \\
\hline
\end{tabular}
\caption{Ablation study results for \textbf{\groupone}. The columns represent the script of the source language. The results are averaged over all target languages. \textbf{Bold} (\udl{underlined}): best (second-best) result.}
\label{table:ablation_group1}
\end{table*}

\begin{table*}[ht]
\setlength{\belowcaptionskip}{-0.3cm}
\scriptsize
\setlength{\tabcolsep}{3pt}
\renewcommand{\arraystretch}{1.2} 
\centering
\begin{tabular}{l|cccc|cccc|cccc|cccc|ccc|}
\toprule
& \multicolumn{4}{c|}{SR-B} & \multicolumn{4}{c|}{Taxi1500} & \multicolumn{4}{c|}{SIB200} & \multicolumn{4}{c|}{NER} & \multicolumn{3}{c|}{POS} \\
& Latn & Hang & Hani & Thai & Latn & Hang & Hani & Thai & Latn & Hang & Hani & Thai & Latn & Hang & Hani & Thai & Latn & Hang & Hani \\
\midrule
Glot500        &  39.4 & 43.0 & 47.1 & 41.7 & 60.0 & 59.2 & \textbf{60.7} & 58.2 & 79.3 & 79.4 & 78.0 & 80.8 & 30.2 & 38.4 & \textbf{48.4} & \textbf{16.4} & \textbf{41.2} & \textbf{39.1} & 54.4 \\
\midrule
MLM            &  45.8 & 54.3 & 54.7 & 51.4 & 58.5 & 60.2 & \udl{60.4} & \udl{60.3} & 80.9 & \udl{80.3} & 79.4 & \udl{81.6} & 28.7 & 37.9 & 46.7 & 7.2  & 29.5 & 27.3 & 54.2 \\
MLM+SEQ        & \udl{ 47.0} & \udl{63.8} & \udl{61.0} & \udl{56.1} & 60.6 & 59.9 & 60.0 & \textbf{61.0} & \udl{81.8} & 79.3 & 79.3 & 80.3 & 30.9 & \textbf{39.3} & 46.5 & 7.7  & 31.4 & 30.9 & \udl{55.4} \\
MLM+TLM        &  45.4 & 55.7 & 57.8 & 54.5 & \textbf{61.4} & \udl{60.6} & 59.5 & 60.0 & \textbf{81.9} & \textbf{81.5} & \udl{79.5} & 81.5 & \udl{31.0} & \udl{39.0} & 47.6 & \udl{9.1 } & 34.6 & \udl{32.1} & 54.8 \\
MLM+SEQ+TLM    & \textbf{ 47.2} & \textbf{70.9} & \textbf{65.8} & \textbf{61.8} & \udl{61.0} & \textbf{61.5} & 59.7 & 60.2 & \udl{81.8} & \textbf{81.5} & \textbf{79.6} & \textbf{81.7} & \textbf{31.9} & 38.6 & \udl{47.8} & 9.0  & \udl{35.8} & 31.8 & \textbf{56.3} \\
\hline
\end{tabular}
\caption{Ablation study results for \textbf{\grouptwo}. The columns represent the script of the source language. The results are averaged over all target languages. \textbf{Bold} (\udl{underlined}): best (second-best) result.}
\label{table:ablation_group2}
\end{table*}

\paragraph{Source languages matter.} We observe that the performance can vary significantly for both groups depending on the
source language used for transfer. This phenomenon occurs due to the script and language similarity between specific source and target languages.
In general, transferring from in-group high-resource languages performs better than transferring from English. Taking the SR-B task, for example,
for \groupone, the best performance is achieved when transferring from Farsi. For the \grouptwo, the best performance is achieved when transferring
from Korean. The text classification tasks generally show less variation in performance, though transferring from Hebrew achieves the worst performance
for \groupone in Taxi1500. For the NER and POS tasks, transferring from Arabic and Hebrew achieves
the best performance for \groupone, while transferring from Chinese achieves the best performance for both tasks in \grouptwo.
Nevertheless, comparing our aligned models against Glot500, the performance generally improves for most source languages.
This indicates that our proposed transliteration-based post-training method effectively improves the alignment between related languages
and further boosts performance when a proper source language is chosen.

\subsection{Ablation Study}\label{subsec:ablation-study}
We perform an ablation study to investigate the impact of different training objectives on the performance of the post-trained aligned models.
Starting from the base Glot500 model, we apply different combinations of the training objectives: masked language modeling (MLM),
sentence-level alignment (SEQ), and token-level alignment (TLM). There are four different combinations in total
(MLM+SEQ is the training objective of \textsc{TransliCo}~\citep{liu2024translico}). Note that we do not consider the variant where
the MLM objective is missing since MLM is important to preserve the language modeling capability. We report the average performance
of using four different source languages across all target languages for each language group in Table~\ref{table:ablation_group1} and
Table~\ref{table:ablation_group2}.

The lone MLM objective already slightly improves the base Glot500 model. We hypothesize this is due to the benefit
of specializing the model to a small group of languages. When the sequence-level alignment (SEQ) objective is included, the performance
is generally further improved compared with the MLM variant, especially for the retrieval task. This is not surprising as the SEQ improves
the sequence-level alignment.
When instead the token-level alignment (TLM) objective is included, there is a slight improvement in the performance for most tasks
compared to the MLM+SEQ objective, except for the retrieval task.
We also observe that text classification tasks show the least variation in performance for most source languages. This is especially the case for SIB200, which seems to be the most straightforward task for the models.

Our whole objective (MLM+SEQ+TLM) generally performs better than the other models by combining sequence
and token-level alignment benefits. For \groupone, though MLM+SEQ+TLM performs on par with the MLM+SEQ objective on NER,
it achieves better performance on POS. Similarly, our complete training objective outperforms MLM+SEQ on both NER and POS for \grouptwo.
When comparing the performance of the full training objective against MLM+TLM, we observe an equal performance across different tasks,
except for the retrieval task, where the full training objective noticeably outperforms MLM+TLM. Even though the SEQ objective is the most critical
for improving the retrieval task, we also observe a performance increase in the retrieval task when TLM is included.
This indicates that our post-training alignment method, with TLM, effectively improves both the sequence and token-level alignment.

\section{Conclusion}\seclabel{sec:conclusion}
In this work, we propose a transliteration-based post-training method that includes both sequence and token-level objectives to improve
the cross-lingual alignment of mPLMs and thus boost their cross-lingual transfer performance. We apply our post-training method
to fine-tune Glot500 on two language groups that share areal features and have extensive lexical overlap. Our extensive experiments
using different source languages show that our aligned models consistently outperform the original Glot500 model.
In particular, our method enhances the alignment and cross-lingual transfer between related languages. We also analyze the impact of different training objectives and show that the sequence and token-level alignment objectives
are both critical for achieving the best performance across different tasks.

\section*{Limitations}
Even though the mPLM is fine-tuned with our method, where the transliterated text is used as an auxiliary input,
the mPLM has only seen the Latin transliterations during its pre-training phase. This can limit the performance of the post-trained aligned models,
especially for languages with complex scripts. An extension of this work could expand the vocabulary to include the subwords
from Latin transliterations as done by \citet{liu2024transmi} before fine-tuning or continually pre-training the model
on the transliterated text so that the models can be more effective in modeling the transliterated data.

We are further limited by the transliteration process, which only partially captures the phonetic and semantic information of the original text,
especially for languages with significantly different scripts from the Latin script. This leads to a loss of information during
the alignment process, which can negatively impact the performance of the post-trained aligned models. Future work could improve
the transliteration process to better capture the linguistic properties of the original text.

Future work could also explore explicit token-level alignment objectives, such as minimizing the L2 distance
between word/subword pair representations as done by \citet{cao2019multilingual}. However, this kind of alignment
cannot be directly applied to many languages in our study. For the \grouptwo group, most language scripts
are logographic or do not use word boundaries, making it challenging to provide an explicit supervision signal
for token-level alignment, as much of this information is lost during the transliteration process.

\section*{Acknowledgments}
This work was funded by Deutsche Forschungsgemeinschaft
(project SCHU 2246/14-1)
and The European Research Council (NonSequeToR, grant \#740516).
We would like to thank Leibniz Supercomputing Center (LRZ) and Munich Center for Machine Learning (MCML) for providing the computational resources.

\bibliography{anthology,anthology_2,custom}

\begin{thebibliography}{42}
\providecommand{\natexlab}[1]{#1}

\bibitem[{Adelani et~al.(2024)Adelani, Liu, Shen, Vassilyev, Alabi, Mao, Gao,
  and Lee}]{adelani-etal-2024-sib}
David Adelani, Hannah Liu, Xiaoyu Shen, Nikita Vassilyev, Jesujoba Alabi, Yanke
  Mao, Haonan Gao, and En-Shiun Lee. 2024.
\newblock \href {https://aclanthology.org/2024.eacl-long.14} {{SIB}-200: A
  simple, inclusive, and big evaluation dataset for topic classification in
  200+ languages and dialects}.
\newblock In \emph{Proceedings of the 18th Conference of the European Chapter
  of the Association for Computational Linguistics (Volume 1: Long Papers)},
  pages 226--245, St. Julian{'}s, Malta. Association for Computational
  Linguistics.

\bibitem[{Amrhein and Sennrich(2020)}]{amrhein-sennrich-2020-romanization}
Chantal Amrhein and Rico Sennrich. 2020.
\newblock \href {https://doi.org/10.18653/v1/2020.findings-emnlp.223} {On
  {R}omanization for model transfer between scripts in neural machine
  translation}.
\newblock In \emph{Findings of the Association for Computational Linguistics:
  EMNLP 2020}, pages 2461--2469, Online. Association for Computational
  Linguistics.

\bibitem[{Anastasopoulos and Neubig(2019)}]{anastasopoulos-neubig-2019-pushing}
Antonios Anastasopoulos and Graham Neubig. 2019.
\newblock \href {https://doi.org/10.18653/v1/D19-1091} {Pushing the limits of
  low-resource morphological inflection}.
\newblock In \emph{Proceedings of the 2019 Conference on Empirical Methods in
  Natural Language Processing and the 9th International Joint Conference on
  Natural Language Processing (EMNLP-IJCNLP)}, pages 984--996, Hong Kong,
  China. Association for Computational Linguistics.

\bibitem[{Artetxe et~al.(2020{\natexlab{a}})Artetxe, Ruder, and
  Yogatama}]{artetxe-etal-2020-cross}
Mikel Artetxe, Sebastian Ruder, and Dani Yogatama. 2020{\natexlab{a}}.
\newblock \href {https://doi.org/10.18653/v1/2020.acl-main.421} {On the
  cross-lingual transferability of monolingual representations}.
\newblock In \emph{Proceedings of the 58th Annual Meeting of the Association
  for Computational Linguistics}, pages 4623--4637, Online. Association for
  Computational Linguistics.

\bibitem[{Artetxe et~al.(2020{\natexlab{b}})Artetxe, Ruder, Yogatama, Labaka,
  and Agirre}]{artetxe-etal-2020-call}
Mikel Artetxe, Sebastian Ruder, Dani Yogatama, Gorka Labaka, and Eneko Agirre.
  2020{\natexlab{b}}.
\newblock \href {https://doi.org/10.18653/v1/2020.acl-main.658} {A call for
  more rigor in unsupervised cross-lingual learning}.
\newblock In \emph{Proceedings of the 58th Annual Meeting of the Association
  for Computational Linguistics}, pages 7375--7388, Online. Association for
  Computational Linguistics.

\bibitem[{Artetxe and Schwenk(2019)}]{artetxe-schwenk-2019-massively}
Mikel Artetxe and Holger Schwenk. 2019.
\newblock \href {https://doi.org/10.1162/tacl_a_00288} {Massively multilingual
  sentence embeddings for zero-shot cross-lingual transfer and beyond}.
\newblock \emph{Transactions of the Association for Computational Linguistics},
  7:597--610.

\bibitem[{Cao et~al.(2020)Cao, Kitaev, and Klein}]{cao2019multilingual}
Steven Cao, Nikita Kitaev, and Dan Klein. 2020.
\newblock \href {https://openreview.net/forum?id=r1xCMyBtPS} {Multilingual
  alignment of contextual word representations}.
\newblock In \emph{8th International Conference on Learning Representations,
  {ICLR} 2020, Addis Ababa, Ethiopia, April 26-30, 2020}. OpenReview.net.

\bibitem[{Chaudhary et~al.(2020)Chaudhary, Raman, Srinivasan, and
  Chen}]{chaudhary2020dict}
Aditi Chaudhary, Karthik Raman, Krishna Srinivasan, and Jiecao Chen. 2020.
\newblock Dict-mlm: Improved multilingual pre-training using bilingual
  dictionaries.
\newblock \emph{arXiv preprint arXiv:2010.12566}.

\bibitem[{Chi et~al.(2021)Chi, Dong, Wei, Yang, Singhal, Wang, Song, Mao,
  Huang, and Zhou}]{chi-etal-2021-infoxlm}
Zewen Chi, Li~Dong, Furu Wei, Nan Yang, Saksham Singhal, Wenhui Wang, Xia Song,
  Xian-Ling Mao, Heyan Huang, and Ming Zhou. 2021.
\newblock \href {https://doi.org/10.18653/v1/2021.naacl-main.280} {{I}nfo{XLM}:
  An information-theoretic framework for cross-lingual language model
  pre-training}.
\newblock In \emph{Proceedings of the 2021 Conference of the North American
  Chapter of the Association for Computational Linguistics: Human Language
  Technologies}, pages 3576--3588, Online. Association for Computational
  Linguistics.

\bibitem[{Conneau et~al.(2020)Conneau, Khandelwal, Goyal, Chaudhary, Wenzek,
  Guzm{\'a}n, Grave, Ott, Zettlemoyer, and
  Stoyanov}]{conneau-etal-2020-unsupervised}
Alexis Conneau, Kartikay Khandelwal, Naman Goyal, Vishrav Chaudhary, Guillaume
  Wenzek, Francisco Guzm{\'a}n, Edouard Grave, Myle Ott, Luke Zettlemoyer, and
  Veselin Stoyanov. 2020.
\newblock \href {https://doi.org/10.18653/v1/2020.acl-main.747} {Unsupervised
  cross-lingual representation learning at scale}.
\newblock In \emph{Proceedings of the 58th Annual Meeting of the Association
  for Computational Linguistics}, pages 8440--8451, Online. Association for
  Computational Linguistics.

\bibitem[{Conneau and Lample(2019)}]{conneau2019cross}
Alexis Conneau and Guillaume Lample. 2019.
\newblock \href
  {https://proceedings.neurips.cc/paper/2019/hash/c04c19c2c2474dbf5f7ac4372c5b9af1-Abstract.html}
  {Cross-lingual language model pretraining}.
\newblock In \emph{Advances in Neural Information Processing Systems 32: Annual
  Conference on Neural Information Processing Systems 2019, NeurIPS 2019,
  December 8-14, 2019, Vancouver, BC, Canada}, pages 7057--7067.

\bibitem[{de~Marneffe et~al.(2021)de~Marneffe, Manning, Nivre, and
  Zeman}]{de-marneffe-etal-2021-universal}
Marie-Catherine de~Marneffe, Christopher~D. Manning, Joakim Nivre, and Daniel
  Zeman. 2021.
\newblock \href {https://doi.org/10.1162/coli_a_00402} {{U}niversal
  {D}ependencies}.
\newblock \emph{Computational Linguistics}, 47(2):255--308.

\bibitem[{Deshpande et~al.(2022)Deshpande, Talukdar, and
  Narasimhan}]{deshpande-etal-2022-bert}
Ameet Deshpande, Partha Talukdar, and Karthik Narasimhan. 2022.
\newblock \href {https://doi.org/10.18653/v1/2022.naacl-main.264} {When is
  {BERT} multilingual? isolating crucial ingredients for cross-lingual
  transfer}.
\newblock In \emph{Proceedings of the 2022 Conference of the North American
  Chapter of the Association for Computational Linguistics: Human Language
  Technologies}, pages 3610--3623, Seattle, United States. Association for
  Computational Linguistics.

\bibitem[{Devlin et~al.(2019)Devlin, Chang, Lee, and
  Toutanova}]{devlin-etal-2019-bert}
Jacob Devlin, Ming-Wei Chang, Kenton Lee, and Kristina Toutanova. 2019.
\newblock \href {https://doi.org/10.18653/v1/N19-1423} {{BERT}: Pre-training of
  deep bidirectional transformers for language understanding}.
\newblock In \emph{Proceedings of the 2019 Conference of the North {A}merican
  Chapter of the Association for Computational Linguistics: Human Language
  Technologies, Volume 1 (Long and Short Papers)}, pages 4171--4186,
  Minneapolis, Minnesota. Association for Computational Linguistics.

\bibitem[{Dyer et~al.(2013)Dyer, Chahuneau, and Smith}]{dyer-etal-2013-simple}
Chris Dyer, Victor Chahuneau, and Noah~A. Smith. 2013.
\newblock \href {https://aclanthology.org/N13-1073} {A simple, fast, and
  effective reparameterization of {IBM} model 2}.
\newblock In \emph{Proceedings of the 2013 Conference of the North {A}merican
  Chapter of the Association for Computational Linguistics: Human Language
  Technologies}, pages 644--648, Atlanta, Georgia. Association for
  Computational Linguistics.

\bibitem[{Gao et~al.(2021)Gao, Yao, and Chen}]{gao-etal-2021-simcse}
Tianyu Gao, Xingcheng Yao, and Danqi Chen. 2021.
\newblock \href {https://doi.org/10.18653/v1/2021.emnlp-main.552} {{S}im{CSE}:
  Simple contrastive learning of sentence embeddings}.
\newblock In \emph{Proceedings of the 2021 Conference on Empirical Methods in
  Natural Language Processing}, pages 6894--6910, Online and Punta Cana,
  Dominican Republic. Association for Computational Linguistics.

\bibitem[{H{\"a}mmerl et~al.(2024)H{\"a}mmerl, Libovick{\'y}, and
  Fraser}]{hammerl2024understanding}
Katharina H{\"a}mmerl, Jind{\v{r}}ich Libovick{\'y}, and Alexander Fraser.
  2024.
\newblock \href {https://doi.org/10.18653/v1/2024.findings-acl.649}
  {Understanding cross-lingual {A}lignment{---}{A} survey}.
\newblock In \emph{Findings of the Association for Computational Linguistics
  ACL 2024}, pages 10922--10943, Bangkok, Thailand and virtual meeting.
  Association for Computational Linguistics.

\bibitem[{Hermjakob et~al.(2018)Hermjakob, May, and
  Knight}]{hermjakob-etal-2018-box}
Ulf Hermjakob, Jonathan May, and Kevin Knight. 2018.
\newblock \href {https://doi.org/10.18653/v1/P18-4003} {Out-of-the-box
  universal {R}omanization tool uroman}.
\newblock In \emph{Proceedings of {ACL} 2018, System Demonstrations}, pages
  13--18, Melbourne, Australia. Association for Computational Linguistics.

\bibitem[{Hu et~al.(2021)Hu, Johnson, Firat, Siddhant, and
  Neubig}]{hu-etal-2021-explicit}
Junjie Hu, Melvin Johnson, Orhan Firat, Aditya Siddhant, and Graham Neubig.
  2021.
\newblock \href {https://doi.org/10.18653/v1/2021.naacl-main.284} {Explicit
  alignment objectives for multilingual bidirectional encoders}.
\newblock In \emph{Proceedings of the 2021 Conference of the North American
  Chapter of the Association for Computational Linguistics: Human Language
  Technologies}, pages 3633--3643, Online. Association for Computational
  Linguistics.

\bibitem[{ImaniGooghari et~al.(2023)ImaniGooghari, Lin, Kargaran, Severini,
  Jalili~Sabet, Kassner, Ma, Schmid, Martins, Yvon, and
  Sch{\"u}tze}]{imanigooghari-etal-2023-glot500}
Ayyoob ImaniGooghari, Peiqin Lin, Amir~Hossein Kargaran, Silvia Severini,
  Masoud Jalili~Sabet, Nora Kassner, Chunlan Ma, Helmut Schmid, Andr{\'e}
  Martins, Fran{\c{c}}ois Yvon, and Hinrich Sch{\"u}tze. 2023.
\newblock \href {https://doi.org/10.18653/v1/2023.acl-long.61} {Glot500:
  Scaling multilingual corpora and language models to 500 languages}.
\newblock In \emph{Proceedings of the 61st Annual Meeting of the Association
  for Computational Linguistics (Volume 1: Long Papers)}, pages 1082--1117,
  Toronto, Canada. Association for Computational Linguistics.

\bibitem[{Khemchandani et~al.(2021)Khemchandani, Mehtani, Patil, Awasthi,
  Talukdar, and Sarawagi}]{khemchandani-etal-2021-exploiting}
Yash Khemchandani, Sarvesh Mehtani, Vaidehi Patil, Abhijeet Awasthi, Partha
  Talukdar, and Sunita Sarawagi. 2021.
\newblock \href {https://doi.org/10.18653/v1/2021.acl-long.105} {Exploiting
  language relatedness for low web-resource language model adaptation: {A}n
  {I}ndic languages study}.
\newblock In \emph{Proceedings of the 59th Annual Meeting of the Association
  for Computational Linguistics and the 11th International Joint Conference on
  Natural Language Processing (Volume 1: Long Papers)}, pages 1312--1323,
  Online. Association for Computational Linguistics.

\bibitem[{Kingma and Ba(2015)}]{ba2015adam}
Diederik~P. Kingma and Jimmy Ba. 2015.
\newblock \href {http://arxiv.org/abs/1412.6980} {Adam: {A} method for
  stochastic optimization}.
\newblock In \emph{3rd International Conference on Learning Representations,
  {ICLR} 2015, San Diego, CA, USA, May 7-9, 2015, Conference Track
  Proceedings}.

\bibitem[{Liu et~al.(2024{\natexlab{a}})Liu, Ma, Ye, and
  Schuetze}]{liu2024translico}
Yihong Liu, Chunlan Ma, Haotian Ye, and Hinrich Schuetze. 2024{\natexlab{a}}.
\newblock \href {https://doi.org/10.18653/v1/2024.acl-long.136} {{T}ransli{C}o:
  A contrastive learning framework to address the script barrier in
  multilingual pretrained language models}.
\newblock In \emph{Proceedings of the 62nd Annual Meeting of the Association
  for Computational Linguistics (Volume 1: Long Papers)}, pages 2476--2499,
  Bangkok, Thailand. Association for Computational Linguistics.

\bibitem[{Liu et~al.(2024{\natexlab{b}})Liu, Ma, Ye, and
  Sch{\"u}tze}]{liu2024transmi}
Yihong Liu, Chunlan Ma, Haotian Ye, and Hinrich Sch{\"u}tze.
  2024{\natexlab{b}}.
\newblock Transmi: A framework to create strong baselines from multilingual
  pretrained language models for transliterated data.
\newblock \emph{arXiv preprint arXiv:2405.09913}.

\bibitem[{Loshchilov and Hutter(2019)}]{loshchilov2018decoupled}
Ilya Loshchilov and Frank Hutter. 2019.
\newblock \href {https://openreview.net/forum?id=Bkg6RiCqY7} {Decoupled weight
  decay regularization}.
\newblock In \emph{7th International Conference on Learning Representations,
  {ICLR} 2019, New Orleans, LA, USA, May 6-9, 2019}. OpenReview.net.

\bibitem[{Ma et~al.(2023)Ma, ImaniGooghari, Ye, Asgari, and
  Sch{\"u}tze}]{ma2023taxi1500}
Chunlan Ma, Ayyoob ImaniGooghari, Haotian Ye, Ehsaneddin Asgari, and Hinrich
  Sch{\"u}tze. 2023.
\newblock Taxi1500: A multilingual dataset for text classification in 1500
  languages.
\newblock \emph{arXiv preprint arXiv:2305.08487}.

\bibitem[{Micikevicius et~al.(2018)Micikevicius, Narang, Alben, Diamos, Elsen,
  Garc{\'{\i}}a, Ginsburg, Houston, Kuchaiev, Venkatesh, and
  Wu}]{micikevicius2018mixed}
Paulius Micikevicius, Sharan Narang, Jonah Alben, Gregory~F. Diamos, Erich
  Elsen, David Garc{\'{\i}}a, Boris Ginsburg, Michael Houston, Oleksii
  Kuchaiev, Ganesh Venkatesh, and Hao Wu. 2018.
\newblock \href {https://openreview.net/forum?id=r1gs9JgRZ} {Mixed precision
  training}.
\newblock In \emph{6th International Conference on Learning Representations,
  {ICLR} 2018, Vancouver, BC, Canada, April 30 - May 3, 2018, Conference Track
  Proceedings}. OpenReview.net.

\bibitem[{Moosa et~al.(2023)Moosa, Akhter, and
  Habib}]{moosa-etal-2023-transliteration}
Ibraheem~Muhammad Moosa, Mahmud~Elahi Akhter, and Ashfia~Binte Habib. 2023.
\newblock \href {https://doi.org/10.18653/v1/2023.findings-eacl.50} {Does
  transliteration help multilingual language modeling?}
\newblock In \emph{Findings of the Association for Computational Linguistics:
  EACL 2023}, pages 670--685, Dubrovnik, Croatia. Association for Computational
  Linguistics.

\bibitem[{Muller et~al.(2021)Muller, Anastasopoulos, Sagot, and
  Seddah}]{muller-etal-2021-unseen}
Benjamin Muller, Antonios Anastasopoulos, Beno{\^\i}t Sagot, and Djam{\'e}
  Seddah. 2021.
\newblock \href {https://doi.org/10.18653/v1/2021.naacl-main.38} {When being
  unseen from m{BERT} is just the beginning: Handling new languages with
  multilingual language models}.
\newblock In \emph{Proceedings of the 2021 Conference of the North American
  Chapter of the Association for Computational Linguistics: Human Language
  Technologies}, pages 448--462, Online. Association for Computational
  Linguistics.

\bibitem[{Murikinati et~al.(2020)Murikinati, Anastasopoulos, and
  Neubig}]{murikinati-etal-2020-transliteration}
Nikitha Murikinati, Antonios Anastasopoulos, and Graham Neubig. 2020.
\newblock \href {https://doi.org/10.18653/v1/2020.sigmorphon-1.22}
  {Transliteration for cross-lingual morphological inflection}.
\newblock In \emph{Proceedings of the 17th SIGMORPHON Workshop on Computational
  Research in Phonetics, Phonology, and Morphology}, pages 189--197, Online.
  Association for Computational Linguistics.

\bibitem[{Pan et~al.(2021)Pan, Hang, Qi, Shah, Potdar, and
  Yu}]{pan-etal-2021-multilingual}
Lin Pan, Chung-Wei Hang, Haode Qi, Abhishek Shah, Saloni Potdar, and Mo~Yu.
  2021.
\newblock \href {https://doi.org/10.18653/v1/2021.naacl-main.20} {Multilingual
  {BERT} post-pretraining alignment}.
\newblock In \emph{Proceedings of the 2021 Conference of the North American
  Chapter of the Association for Computational Linguistics: Human Language
  Technologies}, pages 210--219, Online. Association for Computational
  Linguistics.

\bibitem[{Pan et~al.(2017)Pan, Zhang, May, Nothman, Knight, and
  Ji}]{pan-etal-2017-cross}
Xiaoman Pan, Boliang Zhang, Jonathan May, Joel Nothman, Kevin Knight, and Heng
  Ji. 2017.
\newblock \href {https://doi.org/10.18653/v1/P17-1178} {Cross-lingual name
  tagging and linking for 282 languages}.
\newblock In \emph{Proceedings of the 55th Annual Meeting of the Association
  for Computational Linguistics (Volume 1: Long Papers)}, pages 1946--1958,
  Vancouver, Canada. Association for Computational Linguistics.

\bibitem[{Pires et~al.(2019)Pires, Schlinger, and
  Garrette}]{pires-etal-2019-multilingual}
Telmo Pires, Eva Schlinger, and Dan Garrette. 2019.
\newblock \href {https://doi.org/10.18653/v1/P19-1493} {How multilingual is
  multilingual {BERT}?}
\newblock In \emph{Proceedings of the 57th Annual Meeting of the Association
  for Computational Linguistics}, pages 4996--5001, Florence, Italy.
  Association for Computational Linguistics.

\bibitem[{Purkayastha et~al.(2023)Purkayastha, Ruder, Pfeiffer, Gurevych, and
  Vuli{\'c}}]{purkayastha-etal-2023-romanization}
Sukannya Purkayastha, Sebastian Ruder, Jonas Pfeiffer, Iryna Gurevych, and Ivan
  Vuli{\'c}. 2023.
\newblock \href {https://doi.org/10.18653/v1/2023.findings-emnlp.538}
  {{R}omanization-based large-scale adaptation of multilingual language
  models}.
\newblock In \emph{Findings of the Association for Computational Linguistics:
  EMNLP 2023}, pages 7996--8005, Singapore. Association for Computational
  Linguistics.

\bibitem[{Schuster et~al.(2019)Schuster, Ram, Barzilay, and
  Globerson}]{schuster-etal-2019-cross}
Tal Schuster, Ori Ram, Regina Barzilay, and Amir Globerson. 2019.
\newblock \href {https://doi.org/10.18653/v1/N19-1162} {Cross-lingual alignment
  of contextual word embeddings, with applications to zero-shot dependency
  parsing}.
\newblock In \emph{Proceedings of the 2019 Conference of the North {A}merican
  Chapter of the Association for Computational Linguistics: Human Language
  Technologies, Volume 1 (Long and Short Papers)}, pages 1599--1613,
  Minneapolis, Minnesota. Association for Computational Linguistics.

\bibitem[{Tang et~al.(2022)Tang, Deshpande, and Narasimhan}]{tang2022align}
Henry Tang, Ameet Deshpande, and Karthik Narasimhan. 2022.
\newblock Align-mlm: Word embedding alignment is crucial for multilingual
  pre-training.
\newblock \emph{arXiv preprint arXiv:2211.08547}.

\bibitem[{Versteegh(2001)}]{versteegh2001linguistic}
Kees Versteegh. 2001.
\newblock Linguistic contacts between arabic and other languages.
\newblock \emph{Arabica}, 48(4):470--508.

\bibitem[{Wang et~al.(2020)Wang, Xie, Xu, Yang, Neubig, and
  Carbonell}]{wang2019cross}
Zirui Wang, Jiateng Xie, Ruochen Xu, Yiming Yang, Graham Neubig, and Jaime~G.
  Carbonell. 2020.
\newblock \href {https://openreview.net/forum?id=S1l-C0NtwS} {Cross-lingual
  alignment vs joint training: {A} comparative study and {A} simple unified
  framework}.
\newblock In \emph{8th International Conference on Learning Representations,
  {ICLR} 2020, Addis Ababa, Ethiopia, April 26-30, 2020}. OpenReview.net.

\bibitem[{Wei et~al.(2021)Wei, Weng, Hu, Xing, Yu, and Luo}]{wei2020learning}
Xiangpeng Wei, Rongxiang Weng, Yue Hu, Luxi Xing, Heng Yu, and Weihua Luo.
  2021.
\newblock \href {https://openreview.net/forum?id=Uu1Nw-eeTxJ} {On learning
  universal representations across languages}.
\newblock In \emph{9th International Conference on Learning Representations,
  {ICLR} 2021, Virtual Event, Austria, May 3-7, 2021}. OpenReview.net.

\bibitem[{Wellisch et~al.(1978)Wellisch, Foreman, Breuer, and
  Wilson}]{wellisch1978conversion}
Hans~H Wellisch, Richard Foreman, Lee Breuer, and Robert Wilson. 1978.
\newblock The conversion of scripts, its nature, history, and utilization.

\bibitem[{Wen-Yi and Mimno(2023)}]{wen-yi-mimno-2023-hyperpolyglot}
Andrea~W Wen-Yi and David Mimno. 2023.
\newblock \href {https://doi.org/10.18653/v1/2023.emnlp-main.71} {Hyperpolyglot
  {LLM}s: Cross-lingual interpretability in token embeddings}.
\newblock In \emph{Proceedings of the 2023 Conference on Empirical Methods in
  Natural Language Processing}, pages 1124--1131, Singapore. Association for
  Computational Linguistics.

\bibitem[{Wolf et~al.(2020)Wolf, Debut, Sanh, Chaumond, Delangue, Moi, Cistac,
  Rault, Louf, Funtowicz, Davison, Shleifer, von Platen, Ma, Jernite, Plu, Xu,
  Le~Scao, Gugger, Drame, Lhoest, and Rush}]{wolf-etal-2020-transformers}
Thomas Wolf, Lysandre Debut, Victor Sanh, Julien Chaumond, Clement Delangue,
  Anthony Moi, Pierric Cistac, Tim Rault, Remi Louf, Morgan Funtowicz, Joe
  Davison, Sam Shleifer, Patrick von Platen, Clara Ma, Yacine Jernite, Julien
  Plu, Canwen Xu, Teven Le~Scao, Sylvain Gugger, Mariama Drame, Quentin Lhoest,
  and Alexander Rush. 2020.
\newblock \href {https://doi.org/10.18653/v1/2020.emnlp-demos.6} {Transformers:
  State-of-the-art natural language processing}.
\newblock In \emph{Proceedings of the 2020 Conference on Empirical Methods in
  Natural Language Processing: System Demonstrations}, pages 38--45, Online.
  Association for Computational Linguistics.

\end{thebibliography}

\appendix

\section{Training Details}\seclabel{sec:training-details}
For the MLM objective, we use the normal masking probability of 15\%. We use the AdamW optimizer~\citep{ba2015adam,loshchilov2018decoupled}
with an initial learning rate of $2e-5, \beta_1=0.9, \beta_2=0.999, \epsilon=1e-8$, weight decay of $0.01$, and a linear learning rate scheduler
with no warm-up steps. We use FP16 mixed precision training~\citep{micikevicius2018mixed}. Each batch contains sentence pairs in the original
and Latin scripts, with a maximum sequence length of 512 tokens. We set the per-device batch size to 16, gradient accumulation steps to 8,
and use four NVIDIA A100 80GB GPUs. This leads to an effective batch size of 512. The \groupone model is trained for 2 epochs, while the \grouptwo
model is trained for 8 epochs. We use the HuggingFace Transformers library~\citep{wolf-etal-2020-transformers} for all experiments.
The trainings take around 30 hours for both groups. We store the model checkpoints every 2000 steps.

For the sequence-level contrastive learning objective, we unify all the per-device batch sentence embeddings into a global batch in order to
have a larger amount of negative samples. For the token-level alignment objective, differently from the original TLM, we do not reset the
positional embeddings for the second sentence in a given pair.

\section{Downstream Tasks Fine-Tuning}\seclabel{sec:downstream-tasks}
For the downstream tasks that require fine-tuning, we further fine-tune the aligned models on the training set
of a given source language, select the best checkpoint with early stopping based on the f1 score on the source
language's validation set, and evaluate the macro F1 score on the test sets of the remaining languages in each group.
The results are averaged over five different seeds. All models are fine-tuned on a single GPU.
Unless otherwise stated, the same optimizer and scheduler settings are used as in the post-training alignment phase.

\paragraph{SR-B} We use the same setup as~\citet{imanigooghari-etal-2023-glot500}, by calculating the top-10 retrieval accuracy on
500 parallel sentences from the Bible. This task does not require any fine-tuning, and the retrieval is done by computing the cosine similarity
between the average 8th-layer contextual embeddings of the source and target sentences.

\paragraph{Taxi1500}
A multilingual text classification dataset covering more than 1500 languages
formed by classifying 1077 bible verses into six topics~\citep{ma2023taxi1500}.
The learning rate is set to $1e-5$, and we fine-tune the models for 40 epochs with a batch size of 32 and
a maximum sequence length of 100 tokens.

\paragraph{SIB200}
A multilingual text classification dataset covering more than 200 languages formed by classifying
1004 article sentences into 7 topics~\citep{adelani-etal-2024-sib}.
We use the same fine-tuning setup as the Taxi1500 task, except for the maximum sequence length, which is set to 160 tokens.

\paragraph{NER}
We evaluate the models on the WikiAnn named entity recognition dataset~\citep{pan-etal-2017-cross},
a multilingual dataset consisting of articles annotated with 7 different tags.
We set the learning rate to $2e-5$, the batch size to 32, gradient accumulation steps to 2,
and the maximum sequence length to 256 tokens. We fine-tune the models for 5 epochs.

\paragraph{POS}
We evaluate the models on the Universal Dependencies (UD) v2.11
part-of-speech tagging dataset~\citep{de-marneffe-etal-2021-universal}, a multilingual dataset
consisting of sentences annotated with 17 universal POS tags.
We use the same fine-tuning setup as the NER task, except for the number of epochs, which is set to 10.

\section{Vocabulary Analysis}\seclabel{sec:vocabulary-analysis}
We compare the coverage of the Glot500 vocabulary in the original and transliterated corpora by tokenizing the fine-tuning datasets and
counting the number of \textbf{unique} tokens.
The results are shown in Table~\ref{tab:corpus_comparison}. As expected, the transliterated text is represented
by a smaller part of the vocabulary leading to text being broken down into smaller subwords. This result suggests that the performance
of the post-trained aligned models could be further improved by extending the vocabulary of the pre-trained model based on the transliterated corpus.
\begin{table}[H]
\centering
\footnotesize
\begin{adjustbox}{width=0.48\textwidth}
\begin{tabular}{lcc}
\toprule
\textbf{Language Group} & \textbf{Original Tokens} & \textbf{Transliterated Tokens} \\
\midrule
\textbf{\groupone} & 209K & 125K \\
\textbf{\grouptwo} & 120K & 88K \\
\bottomrule
\end{tabular}
\end{adjustbox}
\caption{Number of \textbf{unique} tokens in the Glot500 vocabulary covered in the original and transliterated corpora for each language group. A smaller set of unique tokens is used after transliterating the corpora into the common Latin script.}
\label{tab:corpus_comparison}
\end{table}
\section{Full Ablation Results}\seclabel{sec:full-ablation-results}
We provide the full results of the other models trained with different combinations of training objectives for both language groups in
Table~\ref{table:ablation_group1_mlm}, Table~\ref{table:ablation_group1_mlm_seq}, Table~\ref{table:ablation_group1_mlm_tlm},
Table~\ref{table:ablation_group2_mlm}, Table~\ref{table:ablation_group2_mlm_seq}, and Table~\ref{table:ablation_group2_mlm_tlm}.

\begin{table*}[ht]
\setlength{\belowcaptionskip}{-0.4cm}
\scriptsize
\setlength{\tabcolsep}{3pt}
\renewcommand{\arraystretch}{1.2} 
\centering
\begin{tabular}{l|cccc|cccc|cccc|cccc|cccc|}
\toprule
& \multicolumn{4}{c|}{SR-B} & \multicolumn{4}{c|}{Taxi1500} & \multicolumn{4}{c|}{SIB200} & \multicolumn{4}{c|}{NER} & \multicolumn{4}{c}{POS} \\
& Latn & Arab & Grek & Hebr & Latn & Arab & Grek & Hebr & Latn & Arab & Grek & Hebr & Latn & Arab & Grek & Hebr & Latn & Arab & Grek & Hebr \\
\midrule
tur\_Latn & 78.6 & \udl{89.4} & 67.2 & 38.6 & \udl{65.1} & \udl{65.1} & 58.9 & 36.2 & \udl{86.8} & 86.2 & 84.8 & 82.9 & \udl{77.2} & 73.1 & 76.5 & 73.9 & \udl{73.2} & 51.1 & 66.5 & 68.7 \\
mlt\_Latn & 76.2 & 78.6 & \udl{80.2} & 47.2 & \udl{61.3} & 56.6 & 59.5 & 35.6 & 84.1 & \udl{85.8} & 84.8 & 83.0 & \udl{75.2} & 73.3 & 71.6 & 74.7 & \udl{81.6} & 62.6 & 77.5 & 75.8 \\
ell\_Grek & 57.2 & \udl{64.2} &  src & 36.2 & \udl{61.5} & 60.5 &  src & 32.4 & \udl{82.9} & 81.3 &  src & 79.9 & 73.8 & 74.1 &  src & \udl{75.0} & \udl{85.3} & 57.0 &  src & 63.0 \\
heb\_Hebr & 25.6 & 28.0 & \udl{32.2} &  src & 43.8 & 43.3 & \udl{45.8} &  src & 78.6 & 77.2 & \udl{79.2} &  src & 50.8 & \udl{60.5} & 55.9 &  src & 65.8 & \udl{70.8} & 57.7 &  src \\
amh\_Ethi & 62.8 & \udl{77.8} & 60.8 & 38.2 & 4.8 & 5.1 & \udl{12.5} & 11.6 & 77.8 & \udl{79.4} & 76.6 & 76.0 & 43.8 & \udl{53.2} & 52.7 & 49.8 & 64.5 & 64.4 & 60.5 & \udl{74.7} \\
ara\_Arab &  - &  - &  - &  - &  - &  - &  - &  - &  - &  - &  - &  - & 58.2 &  src & 59.6 & \udl{66.6} & 62.9 &  src & 59.8 & \udl{77.2} \\
arz\_Arab & 30.4 & 42.4 & \udl{59.4} & 51.2 & 39.5 & 38.4 & \udl{47.8} & 37.3 & \udl{84.4} &  src & 82.9 & 80.3 & 58.9 & \udl{77.6} & 65.3 & 67.4 &  - &  - &  - &  - \\
ary\_Arab & 19.4 & 23.4 & 34.0 & \udl{36.0} & 35.2 & 37.8 & \udl{39.2} & 36.0 & 82.9 & 81.7 & \udl{85.3} & 81.5 &  - &  - &  - &  - &  - &  - &  - &  - \\
arb\_Arab & 12.6 & 22.4 & 30.8 & \udl{33.4} &  - &  - &  - &  - & 83.0 & \udl{85.0} & 83.4 & 79.2 &  - &  - &  - &  - &  - &  - &  - &  - \\
fas\_Arab & \udl{89.8} &  src & 79.2 & 40.8 & \udl{70.1} &  src & 64.6 & 33.8 &  - &  - &  - &  - & 55.6 & \udl{67.8} & 61.7 & 58.0 & 69.6 & 68.6 & 60.2 & \udl{72.5} \\
\midrule
\textbf{Average} & 50.2 & 53.2 & \udl{55.4} & 40.2 & \udl{47.7} & 43.8 & 46.9 & 31.8 & \udl{82.6} & 82.4 & 82.5 & 80.4 & 61.7 & \udl{68.5} & 63.3 & 66.5 & 71.9 & 62.4 & 63.7 & \udl{72.0} \\
\hline
\end{tabular}
\caption{Results for the \textbf{\groupone} model with the MLM objective. Columns represent the script of the transfer source language (denoted with ``src''),
while rows represent the target languages. Results are averaged over 5 random seeds.
For each target language, we \udl{underline} the best source transfer score for each task.}
\label{table:ablation_group1_mlm}
\end{table*}

\begin{table*}[ht]
\setlength{\belowcaptionskip}{-0.4cm}
\scriptsize
\setlength{\tabcolsep}{3pt}
\renewcommand{\arraystretch}{1.2} 
\centering
\begin{tabular}{l|cccc|cccc|cccc|cccc|cccc|}
\toprule
& \multicolumn{4}{c|}{SR-B} & \multicolumn{4}{c|}{Taxi1500} & \multicolumn{4}{c|}{SIB200} & \multicolumn{4}{c|}{NER} & \multicolumn{4}{c}{POS} \\
& Latn & Arab & Grek & Hebr & Latn & Arab & Grek & Hebr & Latn & Arab & Grek & Hebr & Latn & Arab & Grek & Hebr & Latn & Arab & Grek & Hebr \\
\midrule
tur\_Latn & 80.6 & \udl{90.0} & 77.2 & 52.4 & 68.1 & \udl{70.5} & 62.2 & 40.0 & 86.8 & \udl{87.4} & 86.8 & 85.9 & \udl{78.0} & 73.0 & 77.2 & 72.7 & \udl{73.5} & 50.0 & 66.7 & 68.2 \\
mlt\_Latn & 83.2 & \udl{90.2} & 87.8 & 62.8 & \udl{64.1} & 59.2 & 59.2 & 40.3 & 84.8 & 85.2 & \udl{85.4} & 84.3 & 74.5 & 71.3 & 72.1 & \udl{75.0} & \udl{82.4} & 61.9 & 76.8 & 75.9 \\
ell\_Grek & 67.0 & \udl{83.0} &  src & 47.4 & 61.6 & \udl{61.8} &  src & 38.9 & \udl{83.1} & 82.0 &  src & 82.2 & 73.1 & 73.6 &  src & \udl{75.1} & \udl{85.9} & 57.5 &  src & 68.6 \\
heb\_Hebr & 32.0 & 37.4 & \udl{45.4} &  src & 45.9 & \udl{46.0} & 45.9 &  src & 79.8 & 77.5 & \udl{80.1} &  src & 50.4 & \udl{60.6} & 57.0 &  src & 66.6 & \udl{71.1} & 58.8 &  src \\
amh\_Ethi & 57.4 & \udl{71.0} & 57.8 & 46.6 & 4.8 & 6.2 & \udl{14.6} & 13.7 & 76.4 & \udl{79.2} & 78.8 & 77.5 & 41.9 & 52.3 & \udl{54.1} & 49.5 & 65.1 & 63.6 & 63.1 & \udl{74.1} \\
ara\_Arab &  - &  - &  - &  - &  - &  - &  - &  - &  - &  - &  - &  - & 56.5 &  src & 61.9 & \udl{65.8} & 63.9 &  src & 61.7 & \udl{78.0} \\
arz\_Arab & 59.0 & 79.2 & \udl{82.6} & 71.4 & 39.1 & 40.3 & \udl{49.3} & 43.0 & 83.6 &  src & \udl{83.9} & 81.8 & 59.1 & \udl{75.6} & 66.9 & 67.3 &  - &  - &  - &  - \\
ary\_Arab & 52.6 & \udl{68.4} & \udl{68.4} & 68.0 & 36.9 & 38.2 & \udl{39.5} & 38.8 & 82.3 & 81.5 & \udl{84.8} & 82.8 &  - &  - &  - &  - &  - &  - &  - &  - \\
arb\_Arab & 47.6 & 56.0 & \udl{56.2} & 52.2 &  - &  - &  - &  - & 82.6 & 84.0 & \udl{85.0} & 82.6 &  - &  - &  - &  - &  - &  - &  - &  - \\
fas\_Arab & \udl{87.0} &  src & 83.0 & 56.2 & \udl{71.2} &  src & 68.6 & 38.0 &  - &  - &  - &  - & 55.1 & \udl{65.2} & 63.0 & 57.4 & 69.9 & 68.2 & 60.7 & \udl{71.3} \\
\midrule
\textbf{Average} & 62.9 & \udl{71.9} & 69.8 & 57.1 & \udl{49.0} & 46.0 & 48.5 & 36.1 & 82.4 & 82.4 & \udl{83.5} & 82.4 & 61.1 & \udl{67.4} & 64.6 & 66.1 & 72.5 & 62.1 & 64.7 & \udl{72.7} \\
\hline
\end{tabular}
\caption{Results for the \textbf{\groupone} model with the MLM+SEQ objective. Columns represent the script of the transfer source language (denoted with ``src''),
while rows represent the target languages. Results are averaged over 5 random seeds.
For each target language, we \udl{underline} the best source transfer score for each task.}
\label{table:ablation_group1_mlm_seq}
\end{table*}

\begin{table*}[ht]
\setlength{\belowcaptionskip}{-0.4cm}
\scriptsize
\setlength{\tabcolsep}{3pt}
\renewcommand{\arraystretch}{1.2} 
\centering
\begin{tabular}{l|cccc|cccc|cccc|cccc|cccc|}
\toprule
& \multicolumn{4}{c|}{SR-B} & \multicolumn{4}{c|}{Taxi1500} & \multicolumn{4}{c|}{SIB200} & \multicolumn{4}{c|}{NER} & \multicolumn{4}{c}{POS} \\
& Latn & Arab & Grek & Hebr & Latn & Arab & Grek & Hebr & Latn & Arab & Grek & Hebr & Latn & Arab & Grek & Hebr & Latn & Arab & Grek & Hebr \\
\midrule
tur\_Latn & 78.2 & \udl{89.4} & 66.2 & 40.0 & 64.0 & \udl{66.2} & 59.9 & 41.2 & \udl{86.8} & 86.2 & 84.9 & 85.2 & \udl{76.7} & 72.6 & 76.4 & 73.3 & \udl{73.3} & 52.4 & 66.9 & 68.9 \\
mlt\_Latn & 76.8 & 82.0 & \udl{82.2} & 48.0 & \udl{66.8} & 61.5 & 60.2 & 45.0 & \udl{86.3} & 85.1 & 85.7 & 83.6 & \udl{75.8} & 72.0 & 72.6 & 75.6 & \udl{83.1} & 64.0 & 77.3 & 78.7 \\
ell\_Grek & 56.6 & \udl{64.8} &  src & 34.0 & 65.1 & \udl{66.5} &  src & 40.0 & \udl{82.5} & 81.5 &  src & 81.8 & 73.6 & 73.7 &  src & \udl{75.5} & \udl{85.8} & 58.3 &  src & 67.4 \\
heb\_Hebr & 26.2 & 27.8 & \udl{33.2} &  src & \udl{47.1} & 43.9 & 41.4 &  src & 77.6 & 77.6 & \udl{80.7} &  src & 51.9 & \udl{60.4} & 56.6 &  src & 67.5 & \udl{70.8} & 59.2 &  src \\
amh\_Ethi & 60.0 & \udl{78.2} & 57.0 & 37.4 & 7.5 & 8.5 & \udl{17.0} & 10.6 & 77.1 & \udl{77.6} & 76.7 & 76.4 & 46.2 & 51.7 & \udl{54.0} & 46.6 & 65.7 & 63.4 & 62.5 & \udl{75.3} \\
ara\_Arab &  - &  - &  - &  - &  - &  - &  - &  - &  - &  - &  - &  - & 60.1 &  src & 59.3 & \udl{66.8} & 65.7 &  src & 60.0 & \udl{77.9} \\
arz\_Arab & 30.2 & 45.4 & \udl{61.2} & 54.4 & 42.8 & 40.9 & \udl{49.4} & 42.1 & \udl{83.9} &  src & 82.2 & 82.6 & 57.6 & \udl{76.0} & 66.2 & 68.9 &  - &  - &  - &  - \\
ary\_Arab & 21.8 & 27.0 & 35.6 & \udl{39.6} & 38.3 & 37.8 & \udl{39.6} & 39.0 & \udl{83.6} & 81.1 & 82.5 & 81.5 &  - &  - &  - &  - &  - &  - &  - &  - \\
arb\_Arab & 13.8 & 23.6 & 31.2 & \udl{36.2} &  - &  - &  - &  - & 83.1 & 82.9 & \udl{84.0} & 81.6 &  - &  - &  - &  - &  - &  - &  - &  - \\
fas\_Arab & \udl{90.0} &  src & 79.6 & 43.4 & \udl{72.0} &  src & 63.5 & 42.2 &  - &  - &  - &  - & 53.5 & \udl{63.2} & 63.0 & 60.7 & 71.6 & 69.6 & 60.9 & \udl{73.0} \\
\midrule
\textbf{Average} & 50.4 & 54.7 & \udl{55.7} & 41.6 & \udl{50.4} & 46.5 & 47.3 & 37.2 & \udl{82.6} & 81.7 & 82.4 & 81.8 & 61.9 & \udl{67.1} & 64.0 & 66.8 & 73.2 & 63.1 & 64.5 & \udl{73.5} \\
\hline
\end{tabular}
\caption{Results for the \textbf{\groupone} model with the MLM+TLM objective. Columns represent the script of the transfer source language (denoted with ``src''),
while rows represent the target languages. Results are averaged over 5 random seeds.
For each target language, we \udl{underline} the best source transfer score for each task.}
\label{table:ablation_group1_mlm_tlm}
\end{table*}

\begin{table*}[ht]
\setlength{\belowcaptionskip}{-0.4cm}
\scriptsize
\setlength{\tabcolsep}{3pt}
\renewcommand{\arraystretch}{1.2} 
\centering
\begin{tabular}{l|cccc|cccc|cccc|cccc|ccc}
\toprule
& \multicolumn{4}{c|}{SR-B} & \multicolumn{4}{c|}{Taxi1500} & \multicolumn{4}{c|}{SIB200} & \multicolumn{4}{c|}{NER} & \multicolumn{3}{c}{POS} \\
& Latn & Hang & Hani & Thai & Latn & Hang & Hani & Thai & Latn & Hang & Hani & Thai & Latn & Hang & Hani & Thai & Latn & Hang & Hani \\
\midrule
tha\_Thai & 39.6 & \udl{47.8} & 46.4 &  src & 65.2 & 65.1 & \udl{70.6} &  src & \udl{85.5} & 84.3 & 83.8 &  src & 1.9 & 0.4 & \udl{8.7} &  src & 47.1 & 29.2 & \udl{48.2} \\
kor\_Hang & \udl{74.4} &  src & 70.0 & 65.4 & 63.3 &  src & 62.0 & \udl{65.4} & 82.8 &  src & 83.0 & \udl{83.8} & \udl{56.2} &  src & 40.7 & 2.9 & \udl{52.9} &  src & 38.2 \\
yue\_Hani & 38.2 & 55.8 & \udl{82.4} & 67.6 & 68.6 & 66.4 & \udl{70.2} & 69.4 & 86.0 & 85.5 &  src & \udl{88.3} & 16.8 & 33.7 & \udl{70.4} & 11.8 & 20.3 & 30.7 & \udl{78.6} \\
wuu\_Hani &  - &  - &  - &  - &  - &  - &  - &  - &  - &  - &  - &  - & 32.2 & 54.2 & \udl{62.8} & 1.3 &  - &  - &  - \\
zho\_Hani & 46.0 & \udl{49.8} &  src & 44.4 & 64.8 & \udl{67.8} &  src & \udl{67.8} &  - &  - &  - &  - & 17.7 & \udl{33.1} &  src & 11.0 & 20.6 & \udl{33.9} &  src \\
lzh\_Hani & 62.2 & 70.4 & \udl{74.8} & 44.6 & 60.0 & 58.5 & \udl{64.4} & 57.6 &  - &  - &  - &  - & 8.2 & 33.3 & \udl{62.0} & 14.7 & 6.7 & 15.2 & \udl{51.6} \\
lao\_Laoo & 53.6 & \udl{78.4} & 53.8 & 72.2 & 67.9 & 67.4 & \udl{72.6} & 69.3 & 83.0 & 80.8 & 81.2 & \udl{83.9} &  - &  - &  - &  - &  - &  - &  - \\
lhu\_Latn & 6.4 & 8.8 & 7.6 & \udl{10.4} & 20.2 & \udl{31.1} & 21.7 & 29.6 &  - &  - &  - &  - &  - &  - &  - &  - &  - &  - &  - \\
mya\_Mymr & 32.8 & \udl{43.2} & 32.8 & \udl{43.2} & 57.8 & \udl{65.2} & 61.3 & 62.9 & 79.6 & 79.6 & \udl{80.5} & 79.6 & 56.6 & \udl{68.6} & 38.7 & 3.6 &  - &  - &  - \\
bod\_Tibt & 59.8 & \udl{80.4} & 70.2 & 64.0 &  - &  - &  - &  - & 68.6 & 71.4 & 68.5 & \udl{72.1} & 39.7 & 42.0 & \udl{43.2} & 5.5 &  - &  - &  - \\
\midrule
\textbf{Average} & 45.8 & 54.3 & \udl{54.7} & 51.4 & 58.5 & 60.2 & \udl{60.4} & 60.3 & 80.9 & 80.3 & 79.4 & \udl{81.6} & 28.7 & 37.9 & \udl{46.7} & 7.2 & 29.5 & 27.3 & \udl{54.2} \\
\hline
\end{tabular}
\caption{Results for the \textbf{\grouptwo} model with the MLM objective. Columns represent the script of the transfer source language (denoted with ``src''),
while rows represent the target languages. Results are averaged over 5 random seeds.
For each target language, we \udl{underline} the best source transfer score for each task.}
\label{table:ablation_group2_mlm}
\end{table*}

\begin{table*}[ht]
\setlength{\belowcaptionskip}{-0.4cm}
\scriptsize
\setlength{\tabcolsep}{3pt}
\renewcommand{\arraystretch}{1.2} 
\centering
\begin{tabular}{l|cccc|cccc|cccc|cccc|ccc}
\toprule
& \multicolumn{4}{c|}{SR-B} & \multicolumn{4}{c|}{Taxi1500} & \multicolumn{4}{c|}{SIB200} & \multicolumn{4}{c|}{NER} & \multicolumn{3}{c}{POS} \\
& Latn & Hang & Hani & Thai & Latn & Hang & Hani & Thai & Latn & Hang & Hani & Thai & Latn & Hang & Hani & Thai & Latn & Hang & Hani \\
\midrule
tha\_Thai & 49.0 & \udl{75.6} & 55.6 &  src & 66.4 & 67.1 & \udl{71.2} &  src & \udl{86.3} & 83.6 & 84.3 &  src & 2.7 & 1.7 & \udl{8.3} &  src & 47.8 & 27.6 & \udl{48.3} \\
kor\_Hang & 58.6 &  src & \udl{72.2} & 70.8 & 67.3 &  src & 61.6 & \udl{68.3} & 82.9 &  src & 83.3 & \udl{83.4} & \udl{55.8} &  src & 41.2 & 3.6 & \udl{53.1} &  src & 41.8 \\
yue\_Hani & 60.2 & 84.4 & \udl{97.4} & 81.0 & 66.2 & 61.9 & \udl{67.7} & 67.2 & 86.9 & 83.7 &  src & \udl{88.0} & 19.3 & 35.0 & \udl{70.5} & 12.2 & 23.0 & 37.9 & \udl{79.4} \\
wuu\_Hani &  - &  - &  - &  - &  - &  - &  - &  - &  - &  - &  - &  - & 44.2 & 58.6 & \udl{63.9} & 2.0 &  - &  - &  - \\
zho\_Hani & 45.2 & \udl{67.4} &  src & 44.4 & 66.2 & 64.5 &  src & \udl{68.0} &  - &  - &  - &  - & 19.9 & \udl{36.2} &  src & 13.2 & 23.5 & \udl{41.2} &  src \\
lzh\_Hani & 36.6 & 37.2 & \udl{55.8} & 30.0 & \udl{65.0} & 59.8 & 60.1 & 58.4 &  - &  - &  - &  - & 9.3 & 31.4 & \udl{62.4} & 18.8 & 9.6 & 16.8 & \udl{52.1} \\
lao\_Laoo & 55.2 & \udl{81.6} & 60.2 & 72.6 & 70.9 & 66.2 & \udl{72.9} & 67.9 & 82.8 & 80.9 & 82.2 & \udl{83.5} &  - &  - &  - &  - &  - &  - &  - \\
lhu\_Latn & 19.4 & \udl{23.4} & 18.0 & 22.4 & 23.1 & \udl{36.7} & 27.6 & 30.7 &  - &  - &  - &  - &  - &  - &  - &  - &  - &  - &  - \\
mya\_Mymr & 40.2 & \udl{58.0} & 49.0 & 50.6 & 59.5 & 63.4 & 58.5 & \udl{66.3} & \udl{81.8} & 79.6 & 78.0 & 78.7 & 57.2 & \udl{65.6} & 43.2 & 2.4 &  - &  - &  - \\
bod\_Tibt & 58.6 & \udl{83.4} & 79.8 & 77.4 &  - &  - &  - &  - & \udl{70.2} & 68.9 & 68.9 & 68.1 & 38.9 & \udl{46.2} & 36.3 & 1.7 &  - &  - &  - \\
\midrule
\textbf{Average} & 47.0 & \udl{63.8} & 61.0 & 56.1 & 60.6 & 59.9 & 60.0 & \udl{61.0} & \udl{81.8} & 79.3 & 79.3 & 80.3 & 30.9 & 39.3 & \udl{46.5} & 7.7 & 31.4 & 30.9 & \udl{55.4} \\
\hline
\end{tabular}
\caption{Results for the \textbf{\grouptwo} model with the MLM+SEQ objective. Columns represent the script of the transfer source language (denoted with ``src''),
while rows represent the target languages. Results are averaged over 5 random seeds.
For each target language, we \udl{underline} the best source transfer score for each task.}
\label{table:ablation_group2_mlm_seq}
\end{table*}

\begin{table*}[ht]
\setlength{\belowcaptionskip}{-0.4cm}
\scriptsize
\setlength{\tabcolsep}{3pt}
\renewcommand{\arraystretch}{1.2} 
\centering
\begin{tabular}{l|cccc|cccc|cccc|cccc|ccc}
\toprule
& \multicolumn{4}{c|}{SR-B} & \multicolumn{4}{c|}{Taxi1500} & \multicolumn{4}{c|}{SIB200} & \multicolumn{4}{c|}{NER} & \multicolumn{3}{c}{POS} \\
& Latn & Hang & Hani & Thai & Latn & Hang & Hani & Thai & Latn & Hang & Hani & Thai & Latn & Hang & Hani & Thai & Latn & Hang & Hani \\
\midrule
tha\_Thai & 37.6 & \udl{46.8} & 45.6 &  src & 64.6 & 63.7 & \udl{69.4} &  src & \udl{86.5} & 84.5 & 85.0 &  src & 1.7 & 2.2 & \udl{9.4} &  src & \udl{50.4} & 30.6 & 47.4 \\
kor\_Hang & 72.8 &  src & \udl{75.0} & 67.6 & \udl{66.9} &  src & 61.4 & 66.0 & 83.6 &  src & 82.8 & \udl{83.7} & \udl{56.0} &  src & 43.4 & 4.0 & \udl{52.8} &  src & 41.2 \\
yue\_Hani & 43.0 & 67.2 & \udl{88.8} & 76.6 & \udl{69.1} & 62.7 & 66.3 & 65.2 & 87.1 & 86.0 &  src & \udl{89.1} & 19.4 & 34.7 & \udl{70.6} & 13.4 & 27.4 & 38.9 & \udl{79.4} \\
wuu\_Hani &  - &  - &  - &  - &  - &  - &  - &  - &  - &  - &  - &  - & 44.5 & 56.6 & \udl{63.0} & 3.6 &  - &  - &  - \\
zho\_Hani & 45.0 & \udl{49.4} &  src & 44.0 & 68.3 & 63.2 &  src & \udl{69.2} &  - &  - &  - &  - & 19.8 & \udl{34.0} &  src & 12.4 & 30.3 & \udl{43.6} &  src \\
lzh\_Hani & 57.6 & 69.6 & \udl{80.0} & 48.4 & \udl{64.0} & 59.7 & 61.6 & 55.4 &  - &  - &  - &  - & 12.1 & 35.4 & \udl{61.1} & 17.2 & 12.1 & 15.3 & \udl{51.3} \\
lao\_Laoo & 57.2 & \udl{77.4} & 54.2 & 71.8 & 72.2 & 68.9 & \udl{72.6} & 70.4 & 81.9 & 82.5 & 81.4 & \udl{83.6} &  - &  - &  - &  - &  - &  - &  - \\
lhu\_Latn & 8.0 & 10.4 & 9.4 & \udl{13.2} & 22.6 & \udl{38.4} & 25.0 & 31.9 &  - &  - &  - &  - &  - &  - &  - &  - &  - &  - &  - \\
mya\_Mymr & 33.0 & 44.2 & 39.4 & \udl{47.4} & 63.6 & \udl{67.9} & 60.1 & 61.6 & 80.9 & \udl{81.2} & 80.0 & 80.3 & 54.3 & \udl{66.3} & 40.5 & 5.4 &  - &  - &  - \\
bod\_Tibt & 54.6 & \udl{80.6} & 70.0 & 67.2 &  - &  - &  - &  - & 71.2 & \udl{73.5} & 68.5 & 70.7 & 40.4 & 44.2 & \udl{44.9} & 8.0 &  - &  - &  - \\
\midrule
\textbf{Average} & 45.4 & 55.7 & \udl{57.8} & 54.5 & \udl{61.4} & 60.6 & 59.5 & 60.0 & \udl{81.9} & 81.5 & 79.5 & 81.5 & 31.0 & 39.0 & \udl{47.6} & 9.1 & 34.6 & 32.1 & \udl{54.8} \\
\hline
\end{tabular}
\caption{Results for the \textbf{\grouptwo} model with the MLM+TLM objective. Columns represent the script of the transfer source language (denoted with ``src''),
while rows represent the target languages. Results are averaged over 5 random seeds.
For each target language, we \udl{underline} the best source transfer score for each task.}
\label{table:ablation_group2_mlm_tlm}
\end{table*}

\end{document}